\documentclass[sn-mathphys-num]{sn-jnl}

\usepackage{graphicx}%
\usepackage{multirow}%
\usepackage{amsmath,amssymb,amsfonts}%
\usepackage{amsthm}%
\usepackage{mathrsfs}%
\usepackage[title]{appendix}%
\usepackage{xcolor}%
\usepackage{textcomp}%
\usepackage{manyfoot}%
\usepackage{booktabs}%
\usepackage{algorithm}%
\usepackage{algorithmicx}%
\usepackage{algpseudocode}%
\usepackage{listings}%
\usepackage{lineno,soul}
\usepackage{setspace}

\newtheorem{proposition}{Proposition}%
\usepackage{soul}

\usepackage{ulem}
\newcommand{\pkg}[1]{{\fontseries{b}\selectfont #1}}

\let\proglang=\textsf

\newcommand{\bftheta}{{\boldsymbol \theta}}
\newcommand{\bfbeta}{{\boldsymbol \beta}}
\newcommand{\bfSig}{{\bf \Sigma}}
\newcommand{\bfs}{{\bf s}}
\newcommand{\bfx}{{\bf x}}
\newcommand{\bfw}{{\bf w}}

\newcommand{\newTxt}{\textcolor{black}}
\renewcommand{\st}[1]{\unskip}

\begin{document}

\title[Compactly-supported nonstationary kernels for exact Gaussian processes]{Compactly-supported nonstationary kernels for computing exact Gaussian processes on big data}


\author*[1]{\fnm{Mark D.} \sur{Risser}}\email{mdrisser@lbl.gov}

\author[2]{\fnm{Marcus M.} \sur{Noack}} 

\author[3]{\fnm{Hengrui} \sur{Luo}} 

\author[3]{\fnm{Ronald J.} \sur{Pandolfi}} 

\affil[1]{\orgdiv{Climate and Ecosystem Sciences Division}, \orgname{Lawrence Berkeley National Lab}, \orgaddress{\city{Berkeley}, \state{CA}, \country{USA}}}

\affil[2]{\orgdiv{Applied Mathematics and Computational Research Division}, \orgname{Lawrence Berkeley National Lab}, \orgaddress{\city{Berkeley}, \state{CA}, \country{USA}}}

\affil[3]{\orgdiv{Department of Statistics}, \orgname{Rice University}, \orgaddress{\city{Houston}, \state{TX}, \country{USA}}}

\onehalfspacing

\abstract{
The Gaussian process (GP) is a widely used \st{probabilistic machine learning method with implicit uncertainty characterization for stochastic function approximation, stochastic modeling, and analyzing real-world measurements of nonlinear processes.} \newTxt{method for analyzing large-scale data sets, including spatio-temporal measurements of nonlinear processes that are now commonplace in the environmental sciences.}
Traditional implementations of GPs involve stationary kernels (also termed covariance functions) that limit their flexibility, and exact methods for inference that prevent application to data sets with more than about ten thousand points. Modern approaches to address stationarity assumptions generally fail to accommodate large data sets, while all attempts to address scalability focus on approximating the Gaussian likelihood, which can involve subjectivity and lead to inaccuracies. In this work, we explicitly derive an alternative kernel that can discover and encode both sparsity and nonstationarity. We embed the kernel within a fully Bayesian GP model and leverage high-performance computing resources to enable the analysis of massive data sets. We demonstrate the favorable performance of our novel kernel relative to existing exact and approximate GP methods across a variety of synthetic data examples. Furthermore, we conduct space-time prediction based on more than one million measurements of daily maximum temperature and verify that our results outperform state-of-the-art methods in the Earth sciences. More broadly, having access to exact GPs that use ultra-scalable, sparsity-discovering, nonstationary kernels allows GP methods to truly compete with a wide variety of machine learning methods.
 }


\maketitle

\section{Introduction} \label{sec:intro}
Gaussian processes (GPs) are a preeminent framework for stochastic function approximation, statistical modeling of real-world measurements,  non-parametric and nonlinear regression within machine learning (ML), and surrogate modeling. 
GPs are analytically tractable and can be fully specified in terms of a prior mean function and a kernel, also known as a covariance function. Additionally, unlike most ML methods, GPs are ``probabilistic'' in the sense that they naturally include measures of uncertainty in predictions and, when implemented in a Bayesian setting, can furthermore account for and propagate uncertainty in their inferred hyperparameters. GPs have been broadly used across many areas of science, 
including statistical modeling of spatio-temporal data \citep{Cressie1991, CressieWikle}, forward modeling \citep{deisenroth2010efficient,vinogradska2016stability,luo2024hybrid}, uncertainty quantification \citep{tuo2022uncertainty,wang2021inference}, and autonomous experimentation \citep{noack2019kriging,stach2021autonomous,noack2021gaussian,thomas2022autonomous}. 

\newTxt{Formally, the GP is a stochastic process model defined on a given domain $\mathcal{X}$ with the property that any finite-dimensional distribution of random variables indexed in $\mathcal{X}$ is Gaussian. A GP is fully specified by a characterization of its first and second moments, and the second-order properties are easily described by the kernel or covariance function $C(\bfx, \bfx')$ $\forall \bfx, \bfx' \in \mathcal{X}$ (throughout this work, we use ``kernel'' to refer to the covariance function). The kernel $C$ is often defined by choosing an appropriate member from the class of positive semi-definite functions for $\mathcal{X}$. When one can write $C(\bfx, \bfx') = C(||\bfx-\bfx'||)$ (i.e., the kernel is only a function of the separation vector $||\bfx-\bfx'||$), the kernel is said to be ``stationary''; otherwise, the kernel is ``nonstationary''. Ultimately, the choice of kernel is highly impactful for a given implementation of a GP, as it describes the dependence or covariance in the stochastic process for different $\bfx, \bfx' \in \mathcal{X}$ and the differentiability of realizations from the stochastic process.}

\newTxt{While GPs are a widely-used method in many areas of science, }
\st{However, there are two main reasons why GPs are not used more broadly for analyzing modern data sets.}
\newTxt{they retain two properties that limit their broader application to modern, large-scale data sets, particularly in the age of artificial intelligence and machine learning.} 
First, traditional implementations of GPs involve stationary kernels \citep[][]{pilario2020review} that are inflexible for modeling real-world data \citep{duan2018mixed}. The fact that standard stationary kernels are so widely used leads to the general impression that GPs themselves are inflexible, as opposed to a particular implementation of GPs being inflexible. Second, evaluating the Gaussian likelihood for $N$ data points scales with $\mathcal{O}(N^2)$ in memory and $\mathcal{O}(N^3)$ in computation, such that GPs are computationally infeasible for as few as ten thousand data points. Given the fact that modern data sets quite commonly involve millions or even billions of data points, in an ML context, this lack of scalability leads users to gravitate towards neural networks and deep learning methods that are easily applied to big data.

The statistics and ML literature contains a wide variety of methods to address these two primary limitations of GPs. Much work has been done to derive and implement nonstationary kernels, wherein the dependence between two input locations cannot be written as a function of their separation distance \citep[see, e.g.,][]{Sampson1992, Higdon1998, Fuentes2001, Paciorek2006, damianou2013, wilson16}. However, with a few exceptions, these methods are generally difficult to implement for large data sets due to their large numbers of hyperparameters and the need for specialized algorithms for training. These methods can also lead to overfitting
without suitable regularization techniques \citep{dearmon2016g, manzhos2023rectangularization, williams2006gaussian,hrluo_2022e,luo2024multiple}. 
Regarding scalability, a variety of methods have been proposed that derive approximations to the GP likelihood, e.g., local GP experts \citep{cohen2020healing}, predictive process and low-rank methods \citep{Banerjee2008, Cressie2008,hrluo_2019a}, covariance tapering \citep{Furrer2006,Kaufman2008}, structure-exploiting methods \citep{wilson2015kernel}, and Vecchia approximations \citep{vecchia1988estimation, Katzfuss2021,szabo2024vecchia}. We refer the interested reader to \cite{heaton2018case} for a direct comparison of many of these methods on a common data set. These so-called ``approximate GP'' methods are not, in general, agnostic to the underlying kernel, and theoretical properties are, in most cases, derived only for stationary kernels. Furthermore, these methods require subjective choices -- e.g., Vecchia approximations necessitate specifying the number of neighbors to condition on, the ordering of points, and the selection of nearest neighbors -- that may introduce sensitivities in a given analysis.
 
A final limitation of most kernels used in the literature (both stationary and nonstationary) is that they cannot model covariances that are exactly zero. In almost all cases, as the separation between two points increases, a kernel will decay to (but not technically reach) zero, yielding dense covariance matrices.
It is our hypothesis that most (if not all) real-world data sets implicitly include at least some degree of sparsity in the covariance, but typical classes of kernels cannot flexibly model both zero and nonzero covariances. 
Compactly-supported kernels \citep[see, e.g.,][]{Wendland1995, buhmann2001new, genton2001classes, Gneiting2002, melkumyan2009sparse} \textit{can} model both zero and nonzero covariances in a data-driven manner, albeit usually with stationary kernels. Approximate GP methods also ``model'' sparsity, albeit via 
ad hoc choices such as the ordering of points and neighbor selection. The combination of nonstationary and also compactly supported kernels would enable more appropriate modeling of real-world data and yield computational shortcuts for the matrix operations needed to calculate the Gaussian likelihood. In short, the density of the GP prior covariance is not fundamental to the method itself but a result of poorly chosen kernels. Instead of using these standard kernels and then retroactively working on scalability, we propose to give the GP the ability to discover sparsity during the model fitting process. 

\begin{figure}[!t]
\begin{center}
\includegraphics[trim={0 0 0 0mm}, clip, width = \textwidth]{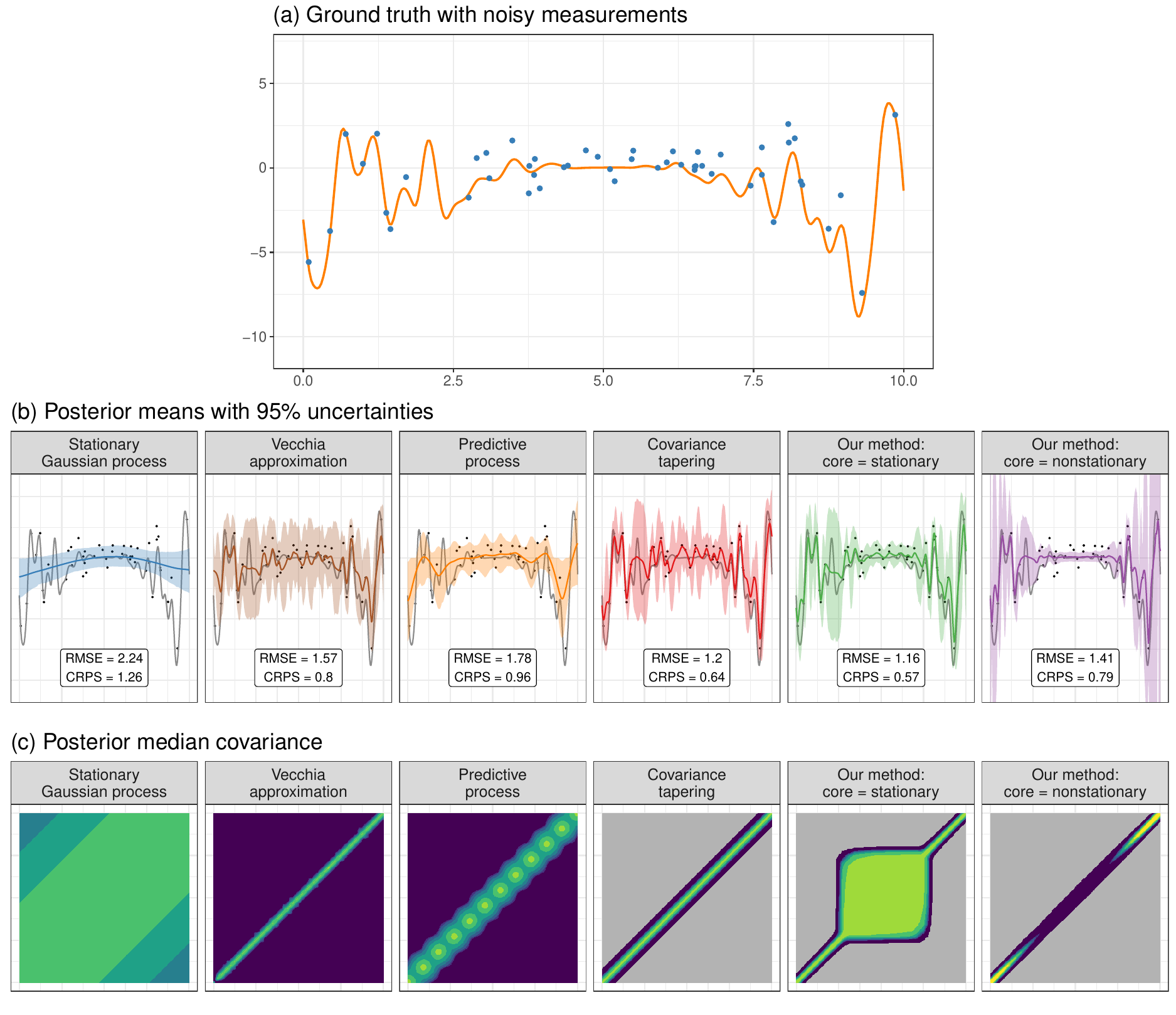}
\caption{Ground truth function over $[0, 10]$ with 50 noisy measurements (panel a.); the synthetic function is both sparse and nonstationary. Panel (b) shows the resulting posterior mean function with 95\% uncertainty intervals for six methods: a stationary Gaussian process with a squared-exponential (non-sparse) kernel, the Vecchia approximation \citep{Katzfuss2021}, the predictive process \citep{Banerjee2008}, covariance tapering \citep{Furrer2006,Kaufman2008}, and our sparsity-discovering kernel combined with both stationary and nonstationary core kernels. Panel (c) shows posterior median covariance estimates (gray areas indicate where the kernel is identically zero). Our kernel has three clear benefits: first, it can discover sparsity in the data while using an exact Gaussian likelihood; second, it can infer the nonstationary structure in the data wherein diagonal elements of the covariance are larger at the endpoints of the domain; third, it outperforms exact and leading approximate methods in terms of root mean square error (RMSE) and continuous rank probability score (CRPS; lower scores for both metrics indicate a better fit). }
\label{Fig_overview}
\end{center}
\end{figure}

In this paper, we introduce a novel 
compactly supported kernel that can discover and encode sparsity, 
yielding ultra-flexible kernels that allow one to apply exact nonstationary GPs to arbitrarily large data sets. A demonstration of our proposed methodology is shown in Figure~\ref{Fig_overview}. Together with prior regularization, the induced sparsity also prevents overfitting and leads to a better interpretation of the data since it preserves the flexibility of a nonstationary kernel.  We also propose a complete Bayesian stochastic model based on the new kernel that accounts for a generic prior mean function and measurement error and allows posterior prediction for unobserved input locations with uncertainty quantification. While high-performance computing (HPC) resources are needed to apply our method to truly massive data sets, we also show how the proposed kernel can be used to conduct fully Bayesian analysis for moderately sized data sets with limited computational resources, e.g., using a personal laptop. The efficacy of our novel kernel is demonstrated using both synthetic data examples and a real-world data set from the Earth sciences. 


\section{Sparsity-discovering nonstationary kernels} \label{sec:kernel_design}

One of the core ideas underpinning this methodology is that real-world data sets almost always involve sparse covariances,
but typical classes of kernels cannot flexibly model zero covariances. Let $y(\bfx)$ be a mean-zero Gaussian process defined on a generic input space $\mathcal{X} \subset \mathbb{R}^d$ with a kernel (also called a covariance function) $C_y(\bfx, \bfx'; \bftheta_y) \equiv Cov\big(y(\bfx), y(\bfx') \big)$ for all $\bfx, \bfx' \in \mathcal{X}$ that describes the covariance between the process $y(\bfx)$ and is known up to a set of hyperparameters $\bftheta_y$. In order to explicitly model sparsity in the stochastic process $y(\cdot)$, we propose modeling the kernel as
\begin{equation*} \label{eq:sparseCov}
    C_y(\bfx, \bfx'; \bftheta_y) = C_\text{core}(\bfx, \bfx'; \bftheta_\text{core}) \times C_\text{sparse}(\bfx, \bfx'; \bftheta_\text{sparse}),
\end{equation*}
where $C_\text{core}$ is a positive definite kernel (stationary or nonstationary), $C_\text{sparse}$ is a sparsity-inducing kernel (defined below in Eq. \ref{eq:C_sparse}), and $\bftheta_y = (\bftheta_\text{core}, \bftheta_\text{sparse})$. While similar in appearance to covariance tapering \citep{Furrer2006,Kaufman2008}, our approach has important distinctions which will be discussed in the next section. So long as $C_\text{core}$ and $C_\text{sparse}$ are positive definite, their product will be positive definite  \citep{horn2012matrix}, yielding a powerful framework for building flexible kernels. 

\subsection{Nonstationary kernels that discover sparsity} \label{sec:sparseKernels}

One way to specify $C_\text{sparse}$ such that it is positive definite and can model sparsity is via a kernel that utilizes sums and products of so-called ``bump functions''
\begin{equation} \label{eq:bumpFcn}
g(\bfx) = \left\{ \begin{array}{ll}
   a \exp\left\{ b \left[ 1 - (1-||\mathbf{x}-\mathbf{h}||^2/r^2)^{-1} \right] \right\}  & \text{if } ||\mathbf{x}-\mathbf{h}|| < r  \\
   0  & \text{else},
\end{array} \right. 
\end{equation}
where $a\geq0$ represents the amplitude, $b>0$ is a shape parameter (see Figure~\ref{Fig_BumpFcns} for a visualization of the effect of $b$), $\mathbf{h} \in \mathbb{R}^d$ is the centroid, and $r>0$ is the radius \citep{Noack2017}. Note that $g(\bfx)$ is smooth and exactly zero for point-to-centroid distances greater than or equal to $r$, which is what enables the covariance matrix to be sparse. We combine these bump functions to yield a sparse kernel
\begin{equation} \label{eq:C_sparse}
    C_\text{sparse}(\bfx, \bfx'; \bftheta_\text{sparse}) = s_0 f_0(\bfx, \bfx'; r_0) + \sum_{i=1}^{n_1} f_i(\bfx)f_i(\bfx').
\end{equation}
In Eq.~\ref{eq:C_sparse}, each $f_i ~\forall i\geq 1,$ is the sum of bump functions $g_{ij}$ from Eq.~\ref{eq:bumpFcn} (each with hyperparameters $\{{\bf h}_{ij}, a_{ij}, b_{ij}, r_{ij}\}$), i.e.,
\begin{equation*} \label{eq:sumOfBumps}
f_i(\bfx) = \sum_{j=1}^{n_2} g_{ij}(\bfx), \hskip4ex i = 1, \dots, n_1,
\end{equation*}
$f_0(\bfx, \bfx'; r_0)$ is a compactly supported stationary kernel, and $s_0$ is a scalar parameter to adjust the relative magnitude of $f_0(\bfx, \bfx; r_0)$ and the amplitudes $\{a_{ij}\}$. We include an additive stationary kernel in Eq.~\ref{eq:C_sparse} because there might otherwise be cases where $C(\bfx, \bfx; \bftheta_\text{sparse}) = 0$; specifying the stationary kernel to be compactly supported ensures that the resulting covariance matrix is still sparse. Among the broad range of possibilities for specifying a compactly supported $f_0$ \citep[see, e.g.,][]{buhmann2001new,genton2001classes,Gneiting2002,melkumyan2009sparse}, we opt to use members of the Wendland family of polynomials \citep{Wendland1995}, e.g.,
\begin{equation*} \label{eq:compSupp}
    f_0(\bfx, \bfx'; r_0)= \left\{ \begin{array}{ll}
    \left(1-\frac{||\bfx-\bfx'||}{r_0}\right)^8\left(35\left(\frac{||\bfx-\bfx'||}{r_0}\right)^3 + 25\left(\frac{||\bfx-\bfx'||}{r_0}\right)^2 + \frac{8||\bfx-\bfx'||}{r_0} + 1\right)
    & \text{if } ||\bfx-\bfx'|| < r_0,  \\
       0  & \text{else.}
    \end{array} \right. 
\end{equation*}
The choice of $f_0$ has implications on the differentiability of the resulting kernel; we return to this point later in the section. Considering $n_1$ and $n_2$ fixed and denoting the hyperparameters of the $j^\text{th}$ bump function $g_{ij}$ of the $i^\text{th}$ component of Eq.~\ref{eq:C_sparse} as $\{{\bf h}_{ij}, a_{ij}, b_{ij}, r_{ij}\}$, the entire vector of hyperparameters for $C_\text{sparse}$ is
$
\bftheta_\text{sparse} = \left( s_0, r_0, \{{\bf h}_{ij}, a_{ij}, b_{ij}, r_{ij} : i = 1, \dots, n_1; j = 1, \dots, n_2 \} \right)
$
which involves a total of $2 + n_1n_2(d+3)$ hyperparameters (recall $d$ is the dimension of the input space, such that ${\bf h} \in \mathcal{X} \subset \mathbb{R}^d$). 
While we suppose that $n_1$ and $n_2$ are fixed \textit{a priori}, these are of course important quantities for a given implementation of the kernel; in a Bayesian framework, we use suitable prior distributions to regularize the sparse kernel, effectively enabling the user to choose arbitrary values of  $n_1$ and $n_2$ such that the data can then determine the optimal number of bump functions (see Section~\ref{subsec:CGPM} for further details). 

\newTxt{Before proceeding, we first verify that $C_\text{sparse}$ as defined in Equation~\ref{eq:C_sparse} is a positive definite function and yields positive definite covariance matrices.}
\begin{proposition}\label{prop:one}
    The kernel $C_\text{sparse}$ is strictly positive definite. Furthermore, $C_y$ is strictly positive definite whenever $C_\text{core}$ is. 
\end{proposition}
\noindent The proof is provided in Appendix~\ref{sec:proof}.



\begin{figure}[!t]
\hskip7ex    (a) Baseline example \hskip6ex (b) Shape$\times 0.1$ \hskip9ex (c) Amplitude$\times2$ \hskip9ex  (d) Radius$\times1.5$
\begin{center}
\begin{turn}{90} 
\hskip3ex Bump functions + sums
\end{turn} 
\includegraphics[trim={0 0 0 6mm}, clip, width = 0.95\textwidth]{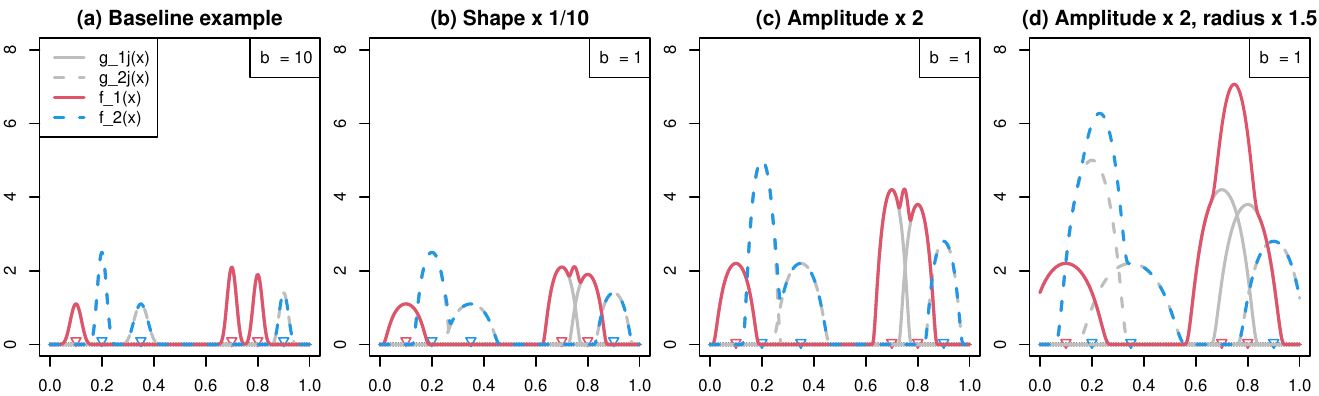} \\[1ex]
\begin{turn}{90} 
\hskip12ex $\sum_{i=1}^2 f_i(x)f_i(x')$
\end{turn} 
\includegraphics[trim={0 10 0 0mm}, clip, width = 0.95\textwidth]{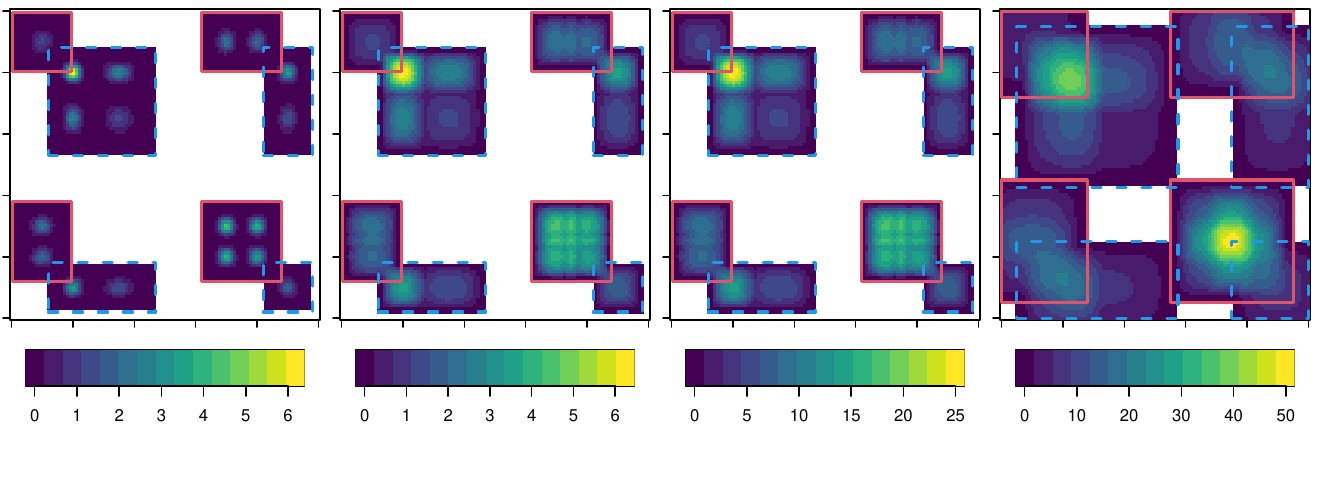}
\begin{turn}{90} 
\hskip12ex $C_\text{sparse}(x, x')$
\end{turn} 
\includegraphics[trim={0 27 0 0mm}, clip, width = 0.95\textwidth]{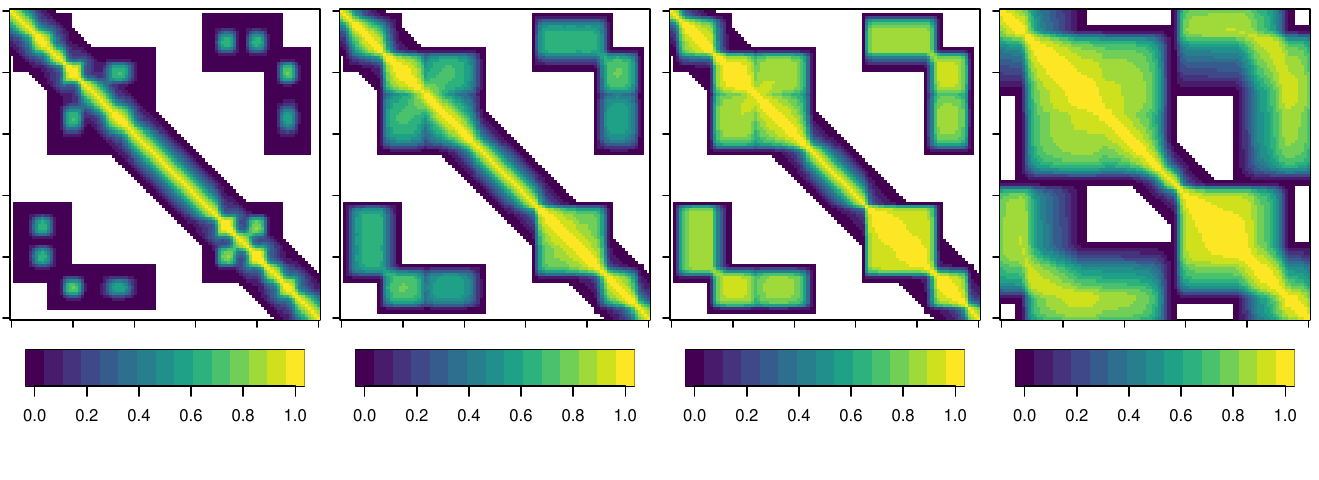}
\caption{Illustration of how the bump functions are used to define the sparse covariance $C_\text{sparse}$ when $n_1=2$ and $n_2=3$ for $N=100$ equally spaced locations on the unit interval $[0,1]$. The top row shows the bump functions $g_{ij}$ and their sum $f_i$ for each $i=1,2$ on the unit interval; the middle row shows image plots of the resulting matrices of entries calculated from the product of bump function sums $\sum_i f_i(x)f_i(x')$; the bottom row shows image plots of $C_\text{sparse}$, i.e., the normalized sum $f_0(x, x'; 0.15) + \sum_i f_i(x)f_i(x')$.
The red (solid) and blue (dashed) lines in the middle row correspond to the effect of bump functions for $i=1$ and $i=2$, respectively.
The different columns show the effects of groups of hyperparameters for fixed bump function centroids: starting from the leftmost column, we consecutively change the shape parameters $b$ (column b.), the amplitudes $a$ (column c.), and the radii (column d.).}
\label{Fig_Example}
\end{center}
\end{figure}

As mentioned previously, \st{the use of} \newTxt{using the} product of covariances has previously been developed under the term ``covariance tapering'' \citep{Furrer2006,Kaufman2008} and by now has almost twenty years of history. Indeed, without the additive cross-product term in Equation~\ref{eq:C_sparse}, the methodology would be essentially identical. However, the primary limitation of covariance tapering is its suboptimal performance when the length-scale of the compactly supported kernel (or ``taper'', i.e., $r_0$) is small relative to the length-scale of the data. In other words, when large-scale or long-range correlations are present in the data, covariance tapering is not the best kernel choice for use within a GP. 
\newTxt{Such long-range correlations are widespread in data sets in the Earth sciences because of so-called ``teleconnections'' (defined as statistical relationships between weather patterns in distant locations), e.g., the El Ni\~no/Southern Oscillation \citep{Philander1985}. }
Importantly, in our formulation, use of the bump functions enables the kernel to capture long-range, off-diagonal correlations (see Figure~\ref{Fig_Example}) such that it maintains strong performance across a wide range of underlying data structures. We compare and contrast our method with covariance tapering directly in Section~\ref{sec:sde2}.

In order to show how the various components of $C_\text{sparse}$ in Eq.~\ref{eq:C_sparse} combine to flexibly model sparsity in the overall kernel $C_y$, we present an illustrative one-dimensional example. We set $n_1=2$ and $n_2 = 3$ and take $\mathcal{X}=[0,1]$ with $N=100$ equally spaced location. Figure~\ref{Fig_Example}(a) shows the bump functions $g_{ij}(x)$ and their sums  $f_i(x)$ using fixed values of the bump function hyperparameters. Starting with the leftmost column of Figure~\ref{Fig_Example}, the 
product of the sums-of-bump-functions $\sum_i f_i(x)f_i(x')$ in the middle row reveals the effect of the bump functions: for a given $i$, the support (i.e., domain of nonzero values) of the bump functions define distance-unrelated parts of the domain that are correlated; furthermore, across the various $i$, a given point in the input space can have a nonzero correlation with different sets of points that does not depend\st{ing} on the distances from this point to these sets. For example, note that both because $f_1(0.17)>0$ and $f_1(0.7)>0$, $C_\text{sparse}(0.17,0.7) > 0$ even though these points are relatively far apart. Furthermore, because $f_2(0.17)>0$ and $f_2(0.9)>0$ (but $f_2(0.7)=0$), we have $C_\text{sparse}(0.17,0.9) > 0$ but $C_\text{sparse}(0.7,0.9) = 0$. The radii $r_{ij}$ specify the size of distance-unrelated regions with nonzero correlations, while the amplitudes $a_{ij}$ and shapes $b_{ij}$ determine the relative magnitude of off-diagonal nonzero correlations and how sharply peaked the correlations are near the bump function centers (Figure~\ref{Fig_Example}b and c). Finally, increasing the radii in Figure~\ref{Fig_Example}(d) predictably decreases the overall sparsity of $C_\text{sparse}$ and (together with relatively small $b$) increases the areas of high correlations. The compactly supported stationary covariance $f_0(x, x'; r_0=0.15)$ serves to ensure that the diagonal elements of $C_\text{sparse}$ are nonzero, with $r_0$ determining the extent to which points very close together are correlated within the sparse kernel.

In summary, each subset of hyperparameters in $C_\text{sparse}$ serves an important purpose for modeling zero and nonzero correlations: the centroids $\{{\bf h}_{ij}\}$ determine portions of the domain that have distance-unrelated (nonstationary) nonzero correlations; the radii $\{r_{ij}\}$ determine the size of the domain subsets that have distance-unrelated nonzero correlations; the shapes $\{b_{ij}\}$ determine how sharply peaked the distance-unrelated nonzero correlations are (sharply peaked for large $b$ and flat for small $b$); the amplitudes $\{a_{ij}\}$ determine the relative magnitude of distance-unrelated nonzero correlations (larger than average $a$ yields larger than average off-diagonal nonzero correlations and vice versa); the compactly supported kernel length-scale $r_0$ determines scales at which nearby points are correlated; and the scaling factor $s_0$ determines the relative magnitude of the compactly supported kernel and the bump functions.

Our specification of $C_\text{sparse}$ implies a nonstationary kernel, irrespective of the choice of $C_\text{core}$: note that we cannot write Eq.~\ref{eq:C_sparse} as a function of the separating vector $\bfx - \bfx'$ {since the bump functions} in Eq.~\ref{eq:bumpFcn} cannot be written as a function of $\bfx - \bfx'$. Our approach explicitly models sparsity directly in the covariance matrix to capture the inherent sparsity of the dependence structure in real-world datasets; alternatively, approximate methods \citep[e.g., inducing points;][]{quinonero2005unifying,hrluo_2019a} focus on approximating the GP with a subset of the data to achieve computational savings. In other words, both classes of methods leverage sparsity to achieve computational savings, but the sparsity implied by our method is data driven and describes the underlying structure in the data set of interest.

Since the bump functions are infinitely differentiable \citep{Noack2017}, we now consider the choice of $f_0$ on the differentiability of $C_y$ at the origin.
\begin{proposition}\label{prop:two}
    The kernel $C_y$ has
    $
    w = \min\{w_0, w_\text{core}\}
    $    
    continuous derivatives at zero, where $f_0$ has $w_0$ continuous derivatives at the origin and $C_\text{core}$ has $w_\text{core}$ continuous derivatives at zero.
\end{proposition}
\noindent The proof is provided in Appendix~\ref{sec:proof}.

\subsection{Modeling choices for the core kernel} \label{sec:Ccore}

In principle, there are no constraints regarding statistical modeling of the core kernel; any positive semi-definite function can be used. In contrast to many approximate methods to scale GPs, our framework allows us to use the full suite of tools that have been developed for kernel modeling, including stationary and nonstationary kernels. For example, even using standard stationary kernels (e.g., the Mat\'ern class) as the core kernel results in the overall kernel being both nonstationary and sparse (since $C_\text{sparse}$ is itself nonstationary). Later in the paper, we utilize parametric families of nonstationary kernels \citep[drawn from, e.g.,][]{Paciorek2006, Risser2015, risser2020bayesian}; in principle, however, one could use any other nonstationary kernel including deep kernels \citep{wilson16} or deep GPs \citep{damianou2013}. In the absence of specific knowledge regarding the form of the core kernel, ``parametric'' nonstationary kernels  \citep{noack2024unifying} such as those derived in \cite{Paciorek2006} and later extended in \cite{risser2020bayesian} have been shown to perform well in practice. See Appendix~\ref{apdx:parnonstat} for more details. 

\subsection{Special cases and infill asymptotics} \label{subsec:specialcases}

Before concluding this section, we briefly describe three special cases of our kernel that can be learned from data via training. First, $C_y$ is sparse and nonstationary by construction, whether or not the core kernel is nonstationary. However, three other cases are possible:
\begin{enumerate}
    \item $C_y$ becomes sparse and stationary if $C_\text{core} = 1$ and all bump functions are set equal to zero (possible if either the radii $\rightarrow0$ or all amplitudes are zero).
    \item $C_y$ becomes non-sparse and stationary if $r_0 \rightarrow \infty$ and $C_\text{core}$ is either (1) chosen to be stationary or (2) collapses to stationary through regularization.
    \item $C_y$ becomes non-sparse and nonstationary if $s_0, r_0 \rightarrow \infty$ and $C_\text{core}$ is nonstationary.
\end{enumerate}
Prior specification and regularization allow the kernel to learn the underlying true state (see Appendix~\ref{sec:sde2}).

\newTxt{Regardless of which case emerges from a given application, we can derive the asymptotic properties our proposed method under the infill framework. As the in-fill design becomes dense, the posterior concentrates on the true kernel parameters, including the bump function radii that dictate where the covariance vanishes. Under this regime, we can establish the consistency of a GP using classical results like Theorem 1 of \cite{van2008rates},
those radii lock onto fixed finite values, so every row of the covariance matrix carries only a bounded number of non-zero entries regardless of sample size. Once this data-driven support has stabilized, the model behaves exactly like a Gaussian process with a conventional, (additive) compactly supported kernels, allowing all classical consistency and efficiency results for such kernels to apply unchanged. We defer a formal statement of this result and the proof to Appendix~\ref{sec:proof}.}

\section{Bayesian stochastic modeling and inference} \label{sec:model}

We now propose a Bayesian framework for statistical modeling with a Gaussian process (Section~\ref{subsec:CGPM}) along with an approach to posterior prediction with uncertainty quantification (Section~\ref{subsec:postPred}). Details regarding the computational requirements of our method are discussed in Section~\ref{subsec:computation}. 

\subsection{Bayesian GP with prior specification, regularization, and training} \label{subsec:CGPM}

Let $\{ z(\bfx): \bfx \in \mathcal{X} \}$ be the observed value of a univariate stochastic process on a given Euclidean domain $\mathcal{X} \subset \mathbb{R}^d$. A general framework for modeling $z(\bfx)$ with a GP is
\begin{equation*} \label{CANONmodel}
z(\bfx) = y(\bfx) + \varepsilon(\bfx), 
\end{equation*}
where $\varepsilon(\cdot)$ is a stochastic error component that is independently distributed as ${N}(0, \tau^2(\bfx))$, with $\varepsilon(\cdot)$ and $y(\cdot)$ are independent. We furthermore assume that the error variance process $\tau^2(\cdot)$ is known up to hyperparameters $\bftheta_z$. The stochastic process $y(\cdot)$ is modeled as a Gaussian process, where the prior mean function $E[y(\bfx)] = {\bf w}(\bfx)^\top \bfbeta$ is linear in a set of covariates or basis functions and the kernel $C_y(\bfx, \bfx'; \bftheta_y) \equiv Cov\big(y(\bfx), y(\bfx') \big)$ is assumed known up to hyperparameters $\bftheta_y$ and describes the covariance between the process $y(\cdot)$ for all $\bfx, \bfx' \in \mathcal{X}$. \newTxt{Note that for spatially-indexed data, $\mathcal{X}  = \mathcal{S} \subset \mathbb{R}^2$ (assuming two dimensional coordinates, e.g., longitude/latitude); for spatio-temporal data where the time dimension is assumed to be continuous, $\mathcal{X} = \mathcal{S} \times \mathcal{T} \subset \mathbb{R}^3$.}

For a fixed and finite set of $N$ input locations $\{{\bf x}_1, ... , {\bf x}_N\}\in \mathcal{X}$, the Gaussian process assumptions imply that the random (observed) vector ${\bf z} = \left[ z({\bf x}_1), ... , z({\bf x}_N) \right]^\top$ has a multivariate Gaussian distribution $p({\bf z} | {\bf y}, \bftheta_z ) = {N}\big({\bf y}, {\bf \Delta}(\bftheta_z)\big)$, where ${\bf \Delta}(\bftheta_z) = diag[\tau^2(\bfx_1), \dots, \tau^2(\bfx_N)]$ and ${\bf y} = \left[ y({\bf x}_1), ... , y({\bf x}_N) \right]^\top$. Conditional on the other hyperparameters in the model, the vector ${\bf y}$ is distributed as $p({\bf y} | \bfbeta, \bftheta_y) = {N}\big({\bf W}\bfbeta, {\bf \Omega}(\bftheta_y) \big)$, where ${\bf W} = [\bfw(\bfx_1)^\top, \dots, \bfw(\bfx_N)^\top]^\top$ and the elements of ${\bf \Omega}(\bftheta_y)$ are $\Omega_{ij} \equiv C_y({\bf x}_i, {\bf x}_j; \bftheta_y)$. To simplify training, one can integrate over the process $y(\cdot)$ to arrive at the marginal distribution of the data:
\begin{equation} \label{Zmarg}
p({\bf z} | \bfbeta, \bftheta) = \int p({\bf z} | {\bf y}, \bftheta_z ) p({\bf y} | \bfbeta, \bftheta_y) d{\bf y} = {N}\big({\bf W}\bfbeta, {\bf \Delta}(\bftheta_z) + {\bf \Omega}(\bftheta_y) \big),
\end{equation}
where $\bftheta = (\bftheta_z, \bftheta_y)$. The kernel for the marginalized process is
\begin{equation} \label{Zcov}
C_z(\bfx, \bfx'; \bftheta) = C_y(\bfx, \bfx'; \bftheta_y) + \tau(\bfx)\tau(\bfx')I_{\{\bfx = \bfx'\}}, \hskip4ex \text{for all } \bfx, \bfx' \in \mathcal{X},
\end{equation}
where $I_{\{\cdot\}}$ is an indicator function. Our approach to modeling $C_y$ was described in Section \ref{sec:kernel_design}.


In a Bayesian setting, one must define prior distributions for the unknown mean and covariance parameters $p(\bfbeta, \bftheta)$; all inference for $\bfbeta$ and $\bftheta$ is then based on the marginalized posterior for these parameters conditional on ${\bf z}$:
\begin{equation} \label{posteriorZ}
p(\bfbeta, \bftheta | {\bf z}) \propto p({\bf z} | \bfbeta, \bftheta) p(\bfbeta,\bftheta).
\end{equation}
Without specific prior knowledge, it is convenient to assume \textit{a priori} that the prior distribution $p(\bfbeta,\bftheta) = p(\bfbeta) p(\bftheta)$ (i.e., $\bfbeta$ and $\bftheta$ are statistically independent) and furthermore that $p(\bfbeta) = N_p({\bf 0}, k{\bf I}_p)$, where $k$ is a large constant (e.g., $k=100^2$) such that the prior for the mean coefficients is noninformative. Regarding prior distributions for $\bftheta_z$ and $\bftheta_y = (\bftheta_\text{core}, \bftheta_\text{sparse})$, proper but noninformative priors will be specified for the elements of $\bftheta_z$.  Specific choices for $\bftheta_\text{core}$ depend of course upon the specific core kernel; see Appendix~\ref{apdx:parnonstat} for more details.
  
For $\bftheta_\text{sparse}$, without loss of generality we fix the bump function shapes $\{b_{ij}\}$ to be 1; otherwise, we specify proper but noninformative priors for the compact stationary length scale and bump function radii:
\[
s_0 \sim U(0,10^5); \hskip5ex  r_0 \sim U(0, D_0); \hskip5ex r_{ij} \sim U(0, D_r), \hskip2ex i = 1, \dots, n_1; j = 1, \dots, n_2
\]
where $U(a,b)$ denotes a uniform distribution on the interval $(a,b)$. Here, $D_0$ and $D_r$ are user-specified upper limits on the local influence of the Wendland and the size of the bump functions' influence, respectively. 
The prior for the bump function locations $\{ {\bf h}_{ij}\}$ is then $h_{ij}^k \sim U(l_k, u_k)$, where the input space $\mathcal{X}\subset \mathbb{R}^d$ (assuming a Euclidean domain) has coordinate-specific lower and upper bounds $\{l_k, u_k : k = 1, \dots, d\}$ (here, $h_{ij}^k$ is the $k^\text{th}$ element of ${\bf h}_{ij}$). Similar priors can be defined for non-Euclidean input spaces. Finally, we propose a simple regularization prior for the bump function amplitudes $\{a_{ij}\}$, where we assume probabilistically that $a_{ij} \sim \text{Bernoulli}(\pi_{ij})$. In other words, the amplitudes are binary random variables that allow individual bump functions $g_{ij}$ to be turned on ($a_{ij}=1$) or off ($a_{ij} = 0$), where $\pi_{ij} = \text{Pr}(a_{ij} =1)$. Lastly, the probabilities $\pi_{ij}$ are assigned uniform priors on the unit interval. In the case that the $ij^\text{th}$ bump function is unnecessary, this formulation will prevent it from having unintended influence on the data-driven sparsity. 

\st{For GPs with trained hyperparameters,}
\newTxt{Despite the fact that the likelihood and $p(\bfbeta)$ are Gaussian, the fact that the kernel parameters $\bftheta$ are unknown means that}
the posterior distribution (\ref{posteriorZ}) is not available in closed form regardless of prior choice \newTxt{$p(\bftheta$). Therefore}, we must resort to Markov chain Monte Carlo (MCMC) methods to conduct inference on $\bfbeta$ and $\bftheta$ \citep{gilks1995markov}. See Appendix~\ref{sec:training} for further details.

\subsection{Posterior prediction with uncertainty quantification} \label{subsec:postPred}

Posterior prediction of the process $y(\cdot)$ for either the observed inputs or a distinct set of $N_P$ unobserved inputs $\{{\bf x}^*_1, ... , {\bf x}^*_{N_P}\}\in \mathcal{X}$ is straightforward given the Gaussian process assumptions used here. In a Bayesian setting, one can conduct either conditional simulation (i.e., conditioning on the data) or unconditional simulation (i.e., without conditioning on the data). Define ${\bf y}_P = ({\bf y}, y({\bf x}^*_1), ... , y({\bf x}^*_{N_P}))$; for conditional simulation the predictive distribution of interest in the Bayesian setting is
\begin{equation*} \label{ppd}
p({\bf y}_P | {\bf z} ) = \int p({\bf y}_P, \bfbeta, \bftheta | {\bf z}) d\bfbeta d\bftheta = \int p({\bf y}_P| \bfbeta, \bftheta, {\bf z}) p(\bfbeta, \bftheta | {\bf z}) d\bfbeta d\bftheta.
\end{equation*}
Based on the Gaussian assumptions, $p({\bf y}_P| \bfbeta, \bftheta, {\bf z})$ can be calculated in closed form as $N\left( {\bf m}_{{\bf y}|{\bf z}}, {\bf C}_{{\bf y}|{\bf z}} \right)$, where
\[
{\bf m}_{{\bf y}|{\bf z}} = {\bf W}_P\bfbeta + {\bf C}_{{\bf y}, {\bf z}} {\bf C}_{{\bf z}}^{-1} ({\bf z} - {\bf W}\bfbeta); \hskip3ex {\bf C}_{{\bf y}|{\bf z}} = {\bf C}_{\bf y} - {\bf C}_{{\bf y}, {\bf z}} {\bf C}_{{\bf z}}^{-1} {\bf C}_{{\bf z}, {\bf y}}.
\]
A Monte Carlo estimate of $p({\bf y}_P | {\bf z} )$ can then be obtained using draws from the posterior, i.e., $p({\bf y}_P | {\bf z} ) \approx \sum_l p({\bf y}_P| \bfbeta^{(l)}, \bftheta^{(l)}, {\bf z})$, where $\{ \bfbeta^{(l)}, \bftheta^{(l)} \}$ are draws from the posterior $p(\bfbeta, \bftheta | {\bf z})$. For unconditional simulation, one can obtain samples of ${\bf y}$ using posterior draws  via ${\bf y}^* = {\bf W}_P{\bfbeta^{(l)}} + {\bf U}^{(l)} {\bf o}$, where ${\bf U}^{(l)}$ is a Cholesky decomposition of ${\bf C}_{\bf y}$ (which is a function of posterior draw ${\bftheta}_l$) and the elements of ${\bf o}$ are iid $N(0,1)$. In other words, this yields ${\bf y}^* \sim N( {\bf W}_P{\bfbeta^{(l)}}, {\bf C}^{(l)}_{\bf y})$.

\subsection{Computational considerations} \label{subsec:computation}

In order to utilize $C_z$ within a Bayesian framework, each iteration of the Markov chain Monte Carlo for the posterior Eq.~\ref{posteriorZ} requires at least one evaluation of the log of the probability density function given in Eq.~\ref{Zmarg} (the log-likelihood), i.e.,
$\log p({\bf z} | \bfbeta, \bftheta) \propto -0.5\log \det({\bf C}_z) - 0.5 ({\bf z} - {\bf W}\bfbeta)^\top {\bf C}_z^{-1} ({\bf z} - {\bf W}\bfbeta)$,
where the elements of ${\bf C}_z$ come from Eq.~\ref{Zcov}. From a computational standpoint, there are three main tasks associated with this evaluation: (1) calculating the covariance matrix ${\bf C}_z$, (2) calculating the log-determinant $\log \det({\bf C}_z)$, and (3) solving the quadratic form $({\bf z} - {\bf W}\bfbeta)^\top {\bf C}_z^{-1} ({\bf z} - {\bf W}\bfbeta)$. The sparsity-discovering nature of our kernel means there are no shortcuts for calculating the covariance matrix for a fixed set of hyperparameters: different elements of ${\bf C}_z$ will be zero vs. nonzero for different hyperparameter configurations. 
\newTxt{In other words, we cannot know ahead of time which elements are zero, and hence must consider all pairs of points when calculating ${\bf C}_z$.}
\st{As such, we must} \newTxt{To get around this bottleneck, instead of calculating a very large covariance matrix on a single node, we} first calculate ${\bf C}_z$ in dense form in batches that are large enough to utilize the chosen hardware well\newTxt{: the ``batches'' represent blocks of the full covariance matrix that can be calculated in parallel (note that the use of batches does not introduce any approximations into the calculation)}. Computationally, the covariance batches are dense but contain many zeros because of our kernel design; therefore, each batch is cast to a sparse format \newTxt{(i.e., we store only the non-zero elements and their row-column position)} on the worker before being transferred back to the host process. There, the full covariance matrix is assembled and exists only in a sparse format leading to minimal RAM requirements. For particularly large datasets, the computation of the batches can be performed by utilizing distributed high-performance-computing architecture. For example, \cite{Noack2023} applied a related kernel to a climate data set of more than 5 million points using HPC resources at the National Energy Research Scientific Computing Center at Lawrence Berkeley National Laboratory.

Steps two and three, namely the log-determinant calculation and the linear-system solve related to the quadratic form can be performed at low computational costs due to the sparse covariance matrix. For example, using the Lanczos method to estimate the log-determinant has a time complexity of $\mathcal{O}(km + k^2)$, where $m$ is the number of nonzero entries in the matrix and $k$ is the number of iterations, which depends on the eigenvalue distribution of the matrix. 
\newTxt{In the biggest data examples in Section~\ref{sec:spattimemod}, we used the \emph{Imate} Python package's Lanczos method with default settings: a normally distributed error in the log-determinant of about 0.1\% (the impact of this on the MCMC's acceptance probability was negligble).}
\st{Further, t} The conjugate gradients (CG) method to solve linear systems has time complexity $\mathcal{O}(mk)$, where $m$ is the number of nonzero elements in the matrix and $k$ is the number of iterations which depend on the condition number of the matrix.
\newTxt{Unfortunately, in practice, since our proposed kernel results in a possibly large range of eigenvalues for the covariance matrix, we found that CG frequently failed to converge.}
\st{For poorly conditioned covariance matrices,} \newTxt{Instead,} we used the minimum residual (MINRES) method which has the same time complexity but \st{is less sensitive to the condition number.} 
\newTxt{is more robust to wildly varying eigenvalues and converges well. We always initialized the MINRES algorithm randomly and with a set error tolerance of $10^{-8}$.}
In practice, we frequently see sparsity numbers of $1/10000$ for large problems of 1 million data points which, together with the proposed kernel design, leads to log-determinant and linear solve compute times in the order of a few second\newTxt{s}. 
\newTxt{Ultimately, in our experience, it has turned out that the combination of the MINRES for linear solves and the Lanczos method for the log-determinant was the only viable option for applying our kernel to data sets with more than one million points.}
For smaller problems ($<100,000$), we take advantage of LU or Cholesky decompositions.  

The fact that our methodology yields no shortcuts for calculating ${\bf C}_z$ is in sharp contrast with approximate methods, e.g., the nearest neighbor Gaussian process (NNGP) proposed in \cite{datta2016hierarchical} and the Vecchia approximations proposed in \cite{Katzfuss2021}. First, the NNGP and Vecchia approximations imply a sparse \textit{precision} matrix (i.e., inverse covariance), while our method implies a sparse \textit{covariance}. Second, the implicit conditioning structure of each method implies that for a given ordering of points and method for choosing nearest neighbors \cite[see choices C1-C5 given in Section 2.3 of][]{Katzfuss2021}, some elements of the precision matrix will always be nonzero, and other elements will always be exactly zero. The fixed nonzero elements are known ahead of time (again, based on specific choices made for ordering of points and selecting neighbors), such that likelihood evaluations for NNGP and Vecchia never involve dense matrices. 

\begin{figure}
    \centering
    \includegraphics[width=\textwidth]{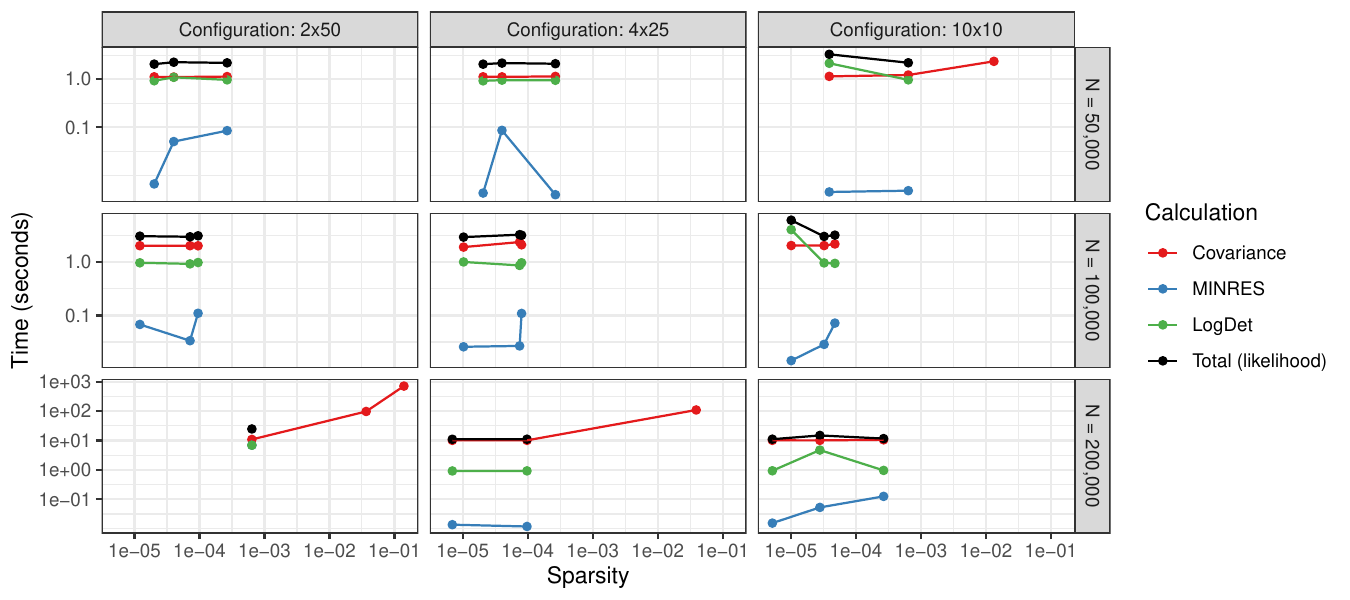}
    \caption{Runtime analysis of the computational cost associated with the proposed methodology. We show the compute time needed for three calculations: (1) the covariance matrix ${\bf C}_z$; (2) the log-determinant using the Lanczos method; and (3) the MINRES linear system solve. Compute time is shown across different combinations of the sample size (rows), the $n_1 \times n_2$ configuration (columns), and sparsity level ($x$-axis). }
    \label{fig:comptime}
\end{figure}

\newTxt{To explore a more detailed runtime analysis of the computational cost associated with our method, we conducted a test using synthetic data across different sample sizes ($N \in \{50000, 100000, 200000\}$), choices of $n_1$ and $n_2$ ($n1\times n2 \in \{2\times 50, 4\times 25, 10\times 10\}$), and sparsity level (ranging from 1e-05 to 1e-01). For each of these combinations, we separately tallied the time needed to calculate the covariance ${\bf C}_z$, the log-determinant calculation using the Lanczos method, and the MINRES linear-system solve; see Figure~\ref{fig:comptime}.
The covariance compute times are based on using 16 GPUs to parallelize across covariance batches; the Lanczos and MINRES times correspond to a single GPU.
Generally speaking, the total time needed to evaluate the likelihood ranges from approximately 2 seconds for $N=50 000$ up to about 10 seconds for $N=200 000$. The slowest operations are calculation of the covariance and the log-determinant, while the MINRES solve is very fast regardless of sample size, $n_1 \times n_2$ configuration, and sparsity. As we would expect, the computational time for calculating the covariance is unaffected by the sparsity: by definition, the covariance calculation requires querying all pairs of points to determine which entries are zero versus non-zero. The computational time is also largely unaffected by the $n_1 \times n_2$ configuration: for a given sample size/sparsity combination, the compute time is roughly the same across $2\times50$, $4\times25$, and $10\times10$. Interestingly, the MINRES and log-determinant calculation are somewhat insensitive to the sparsity, at least in cases where the fraction of nonzero entries is less than 1e-03.
}

\newTxt{In summary, we reiterate that}
\st{At the end of the day,} our methodology is still an exact Gaussian process: \newTxt{the implication is some the computational bottlenecks associated with exact GPs cannot be avoided.} \st{some of the usual challenges associated with computation for a GP cannot be avoided.} 
\newTxt{We make no claims about the computational efficiency of our method relative to approximate GP methods like Vecchia for moderately sized data sets. However, the benefits are clear: we avoid any approximations or alternative formulations (e.g., assuming sparse precision matrices or autoregressive structure) while developing a highly flexible kernel that, when combined with HPC resources, can be used for fully Bayesian analysis of large data sets, on the order of millions of data points. Furthermore, we also gain}
\st{However, what we gain is} substantially less subjectivity: no \textit{a priori} choices need to be made regarding, e.g., the number of neighbors to condition on, the ordering of points, and the selection of nearest neighbors.

\section{Synthetic data examples} \label{sec:synthetic}

We now describe two synthetic data examples to demonstrate the performance of our proposed kernel. The first evaulates performance under mis-specification of the number of bump functions and the efficacy of our proposed regularization priors when too many bump functions are permitted (Section~\ref{sec:sde1}). The second compares our proposed methodology with competing methods across a variety of ground-truth data states (Section~\ref{sec:sde2}). In Section~\ref{sec:sde2}, we assess performance quality using two metrics: root mean square error (RMSE) for the test set, which assesses the pointwise error in posterior mean predictions; and the continuous rank probability score \citep[CRPS;][]{Gneiting2007} for the test set, which (unlike the RMSE) assesses both precision and sharpness of predictions. For example, the CRPS rewards a bad prediction with large uncertainty over a bad prediction with small uncertainty. Note that the reported CRPS scores are calculated for each test data point and then averaged across all test data.

\subsection{Mis-specified bump functions} \label{sec:sde1}

First, we conduct a test to assess the performance of our proposed kernel when the number of bump functions is mis-specified. For a real-world data application, one of course does not know how many bump functions to specify, which is a requisite \textit{a priori} choice that must be made when using our methodology. The goal of this subsection is to provide guidance for making this decision.

\begin{figure}[!t]
\begin{center}
\includegraphics[trim={0 0 0 0mm}, clip, width = 0.75\textwidth]{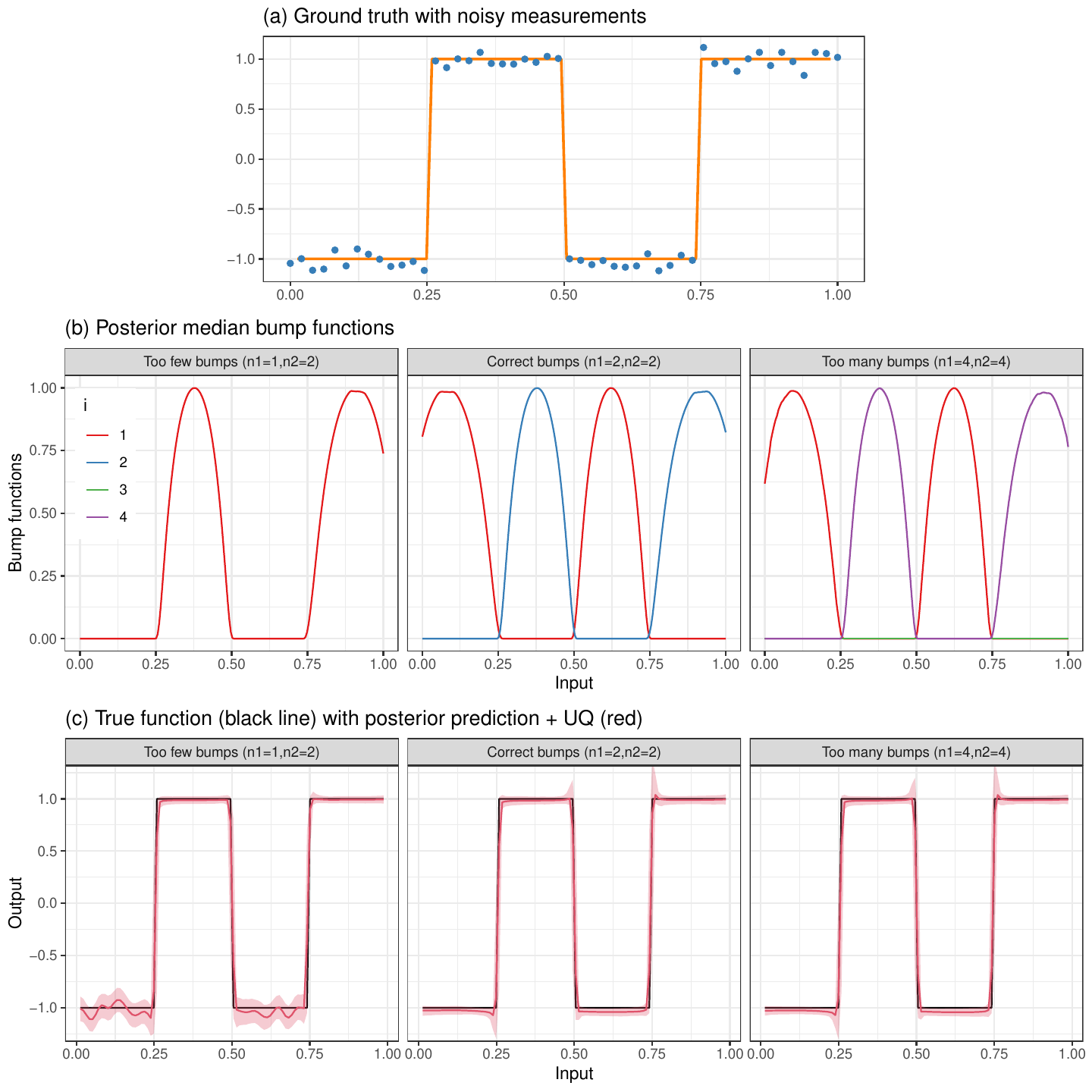}
\caption{Results for the synthetic data test described in Sections~\ref{sec:sde1}. The ground truth function and noisy measurements are shown in panel (a); panels (b) and (c) show the best estimates of the bump functions (posterior medians) and posterior predictions with uncertainty (posterior mean and 90\% credible intervals), respectively. Results are shown when the kernel has too few bump functions (left), the correct number of bump functions (center), and too many bump functions (right).
}
\label{Fig_sde_41}
\end{center}
\end{figure}

The synthetic data test proceeds as follows. First, we suppose the ground truth function is a piecewise-constant function on the unit interval, defined as
\[
y(x) = \left\{ \begin{array}{cc}
    -1 & \text{if } x \leq 0.25 \text{ or } 0.5 < x \leq 0.75  \\
    1 & \text{if } 0.25 < x \leq 0.5 \text{ or } x > 0.75,
\end{array}\right.
\]
and we have noisy measurements from this function for training a GP (see Figure~\ref{Fig_sde_41}a.). The form of this ground truth function suggests that there should be two sets of two bump functions: $y(x)$ is perfectly correlated in $(0,0.25]$ and $(0.5,0.75]$; furthermore, $y(x)$ is perfectly correlated in $(0.25,0.5]$ and $(0.75,1)$. In other words, $n_1=2$ and $n_2=2$. We then fit our proposed kernel to the noisy data assuming we know nothing about its ground truth, under three configurations of bump functions: too few bump functions, where we set $n_1 = 1$ and $n_2=2$; the right number of bump functions, using $n_1 = 2$ and $n_2=2$; too many bump functions, using $n_1 = 4$ and $n_2=4$.
Notice that although the second configuration is the only one with the correct number of bump functions, in all cases we infer the bump functions' characteristics (positions and radii) directly from the data. When fitting, we suppose the core kernel is a stationary Mat\'ern with smoothness $\nu=2.5$.

Results are shown in Figure~\ref{Fig_sde_41}(b) and (c), where we show both the posterior median estimates of the bump functions and the posterior predictions of $y(\cdot)$ with uncertainty, respectively. First, consider the bump function estimates in Figure~\ref{Fig_sde_41}(b). When there are too few bump functions, the training correctly identifies the position and size of the bump functions for part of the domain, i.e., that $(0.25,0.5]$ is correlated with $(0.75,1)$, while the kernel does not have the flexibility to capture the distance-unrelated correlations in $(0,0.25]$ and $(0.5, 0.75]$. As such, the predictions capture the ground truth function with very low uncertainty in $(0.25,0.5]$ and $(0.75,1)$ but revert to the sparse stationary kernel in the other parts of the domain -- effectively over-fitting the data in $(0,0.25]$ and $(0.5, 0.75]$. When there are the correct number of bump functions, the training correctly identifies where these bump functions should go and how large they should be; consequently, posterior predictions are correctly calibrated over the entire unit interval. Finally, when there are too many bump functions, the algorithm correctly determines both the size and position of the bump functions, but also that only four of the 16 total bump functions should be activated. The other 12 are appropriately zeroed out via regularization to ensure that we avoid both under- and over-fitting. Posterior predictions are effectively identical to those generated using the correct number of bump functions.

In summary, this example verifies two important points: (1) our proposed sparse kernel can both identify where and how big the bump functions should be, and (2) the regularization method for the bump functions described in Section~\ref{subsec:CGPM} effectively eliminates the effect of any bump functions that may be unnecessary. In other words, for a real data example, we recommend specifying more bump functions than one might think is truly necessary (i.e., err on the side of making $n_1$ and $n_2$ large). Of course, more bump functions means more hyperparameters to learn during training, so in practice one must maintain a balance between too many hyperparameters and choosing $n_1$ and $n_2$ to be ``large enough.'' 

\subsection{Comparison with competing methods} \label{sec:sde2}


Next, we generate data from a variety of underlying data generating mechanisms and compare our approach with competing methods. The data generating mechanisms are designed to span different combinations of sparsity and nonstationarity that may be present in real-world data sets. Specifically, we generate data from mean-zero multivariate Gaussian distributions where the ``true'' covariance is defined as follows:
\begin{enumerate}
    \item[] S1: Non-sparse and stationary: $C_y$ 
    is the isotropic Mat\'ern kernel. 
    \item[] S2: Sparse and stationary: $C_y(\bfx, \bfx') = \sigma^2 f_0(\bfx,\bfx'; r_0)$, where $f_0$ is as in Eq.~\ref{eq:compSupp}.
    \item[] S3: Sparse and nonstationary: $C_y(\bfx, \bfx') = \sigma(\bfx)\sigma(\bfx') f_0(\bfx,\bfx'; r_0)$, with input-dependent $\sigma(\cdot)$. 
    \item[] S4: Non-sparse and nonstationary: here we take $C_y$ to be the nonstationary kernel described in \cite{risser2020bayesian}.
\end{enumerate}
Here we use S1-S4 to denote different ``scenarios''; see Appendix~\ref{appdx:sde} for more details on the data generating mechanisms, specific hyperparameters used, and sample draws from the true parent distribution. 
In each case, the synthetic data drawn from the parent distribution represents a ground truth function; we can therefore generate $N_\text{train}=50$ (randomly sampled) and $N_\text{test}=300$ (regularly spaced) data points for validation. 
\st{Each training data set will include the addition of white noise that is approximately 10\% of the signal variance.}
\newTxt{We also explore the effect of how large the noise is relative to the signal, and add white noise to the ground truth with a variance that is 5\% (signal-to-noise ratio or SNR of 20), 10\% (SNR = 10), and 20\% (SNR = 5) of the signal variance.}
For each scenario, we generate $N_\text{rep}=50$ replicates or draws from the ground truth distribution; note that the locations of the training data are different across each replicate.

For each scenario and each replicate, we fit a series of constant-mean GP models to the synthetic data, summarized in Table~\ref{tab:simstudy} and mirroring the example given in Figure~\ref{Fig_overview}.  
\newTxt{As a baseline, we fit a stationary Gaussian process with a Mat\'ern kernel, denoted Model 1 or ``M1''.}
\st{a stationary Gaussian process with a Mat\'ern kernel (Model 1 or ``M1'');}
\newTxt{To compare against this baseline, we consider a suite of GP methods:}
the Vecchia approximation \citep[M2,][]{Katzfuss2021}; the predictive process \citep[M3,][]{Banerjee2008}; covariance tapering \citep[M4,][]{Kaufman2008}; our proposed approach, using the kernel proposed in Section~\ref{sec:kernel_design} with a stationary Mat\'ern core kernel (M5); and as in M5, but with a nonstationary core kernel defined in Equation~\ref{PScov} (M6). The Vecchia approximation uses 15 nearest-neighbors; the predictive process uses 12 knots ($12/50=24\%$ of the data size); and the prior for the covariance taper limits the nonzero correlations to be within $\pm1$ unit ($10\%$ of the size of the input space) of each data point. In all cases, the proposed covariance kernel is used within a fully Bayesian model where all hyperparameters are trained separately for each replicate. 

In principle, M1-M4 could involve nonstationary kernels; however, we maintain stationary kernels for two reasons. First, M1 is primarily used here as a reference for what one can obtain with a ``default'' (stationary and exact) GP when working with smaller data sets; the whole point of this methodology is that M1 will become unusable for much larger data sets. Second, all of the underlying theory for Vecchia approximations, predictive process, and covariance tapering has been derived for stationary kernels. For example, one can use the Vecchia framework with a nonstationary kernel \citep[see, e.g.,][]{risser2020bayesian}; however, without further theoretical developments there are no guarantees that the subjective choices one must make for the Vecchia approximation (size of conditioning sets, ordering of points, selection of nearest neighbors) will result in an appropriate approximation for a nonstationary kernel. 
Our implementation of M1 and M2 involve standard, off-the-shelf implementations of these methods as enabled by the \pkg{BayesNSGP} software package for \proglang{R} \citep[][]{R_BayesNSGP}. We implement our own versions of M3 and M4, since existing software \citep[e.g.,][]{Finley2007} cannot be used for one-dimensional inputs.  For M5 and M6, the main user-specified feature is determining how many bump functions are allowed: here, we set  $n_1=4$ and $n_2=4$). In order to ease comparison, we show CRPS \st{and RMSE} for M2-M6 relative to corresponding quantities from M1: relative scores less than one indicate \st{RMSE and} CRPS is smaller (and therefore better) for the model of interest versus a stationary GP. 

\begin{table}[!t]
\begin{center}
{\small 
\begin{tabular}{clcccl}
\hline\noalign{\smallskip}
\multicolumn{2}{c}{\textbf{Data generating mechanisms} } &&& \multicolumn{2}{c}{\textbf{Kernels within the Gaussian process} } \\ 
\textit{Label} & \multicolumn{1}{c}{\textit{Description}} &&&   \textit{Label} & \multicolumn{1}{c}{\textit{Description}} \\ 
\hline\noalign{\smallskip} 
S1 & Non-sparse, stationary          &&& M1 & Stationary Mat\'ern kernel \\
S2 & Sparse, stationary          &&& M2 & Vecchia approximation with Mat\'ern \\
S3 & Sparse, nonstationary          &&& M3 & Predictive process \\
S4 & Non-sparse, nonstationary          &&& M4 & Covariance tapering \\
 &          &&& M5 & Sparsity-discovering kernel with stationary core \\
 &           &&& M6 & Sparsity-discovering kernel with nonstationary core\\
\noalign{\smallskip}\hline
\end{tabular}
}
\end{center}
\caption{Summary of five data-generating mechanisms (or ``scenarios'') used to generate synthetic data and six kernels used within a Gaussian process to fit each of $N_\text{rep}=50$ repeated draws from each scenario.}
\label{tab:simstudy}
\end{table}

\begin{figure}[!t]
\begin{center}
\includegraphics[trim={0 0 0 0mm}, clip, width = \textwidth]{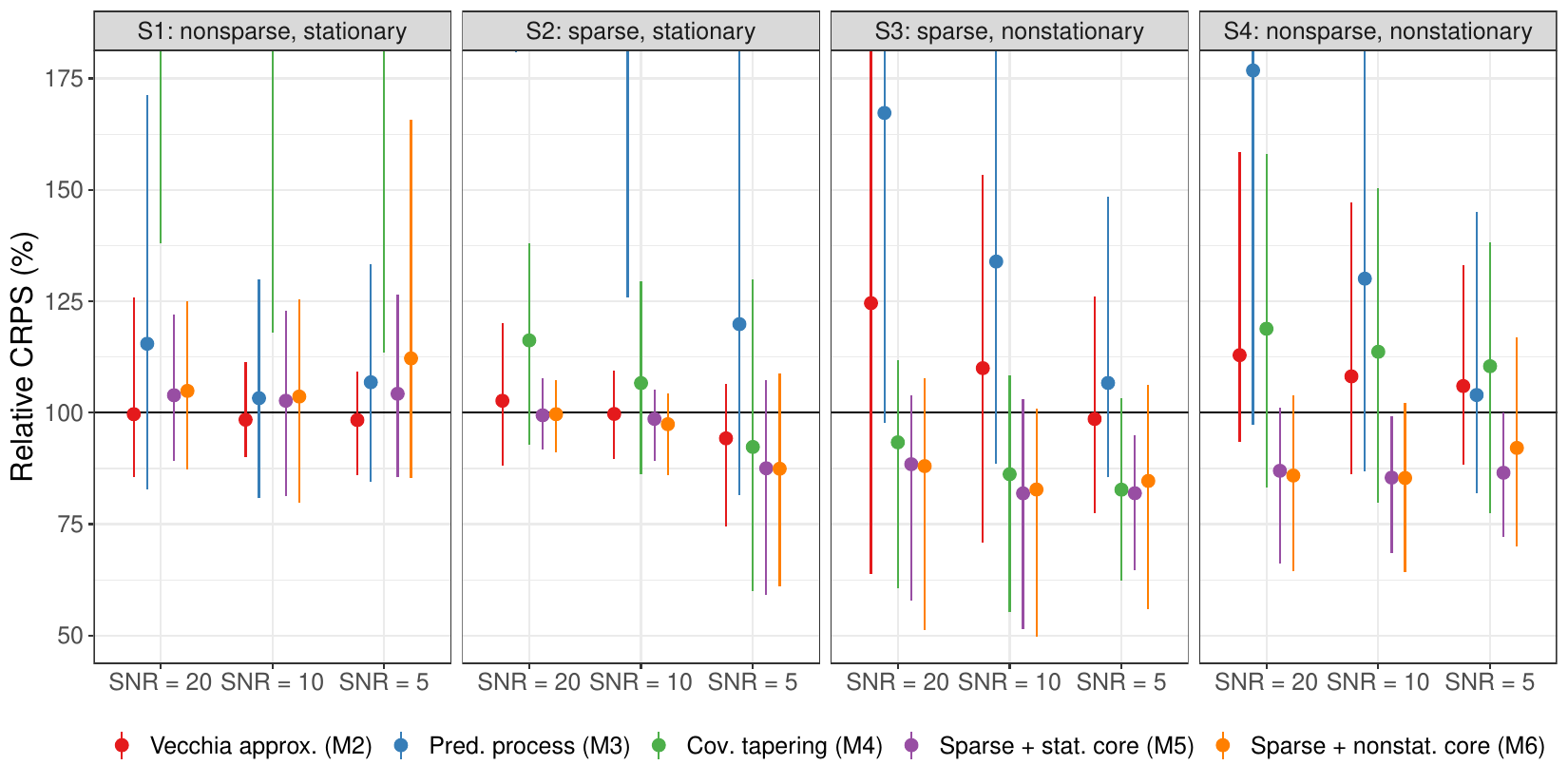}
\caption{Results for the synthetic data tests described in Section~\ref{sec:sde2}. The ground truth ``scenarios'' are \st{labeled on} \newTxt{denoted with different colors, while the different signal-to-noise (SNR) considered are shown on} the $x$-axis. We show the mean and central 90\% of continuous rank probability scores \citep[CRPS;][]{Gneiting2007} \st{and root mean square error (RMSE)} across 50 synthetic data sets for five competing methods relative to fitting a Gaussian process with a stationary Mat\'ern kernel (M1)\newTxt{; relative scores less than 100\% indicate better performance relative to M1}. Scores are tallied for each of the data generating mechanisms describe in Table~\ref{tab:simstudy}.
}
\label{Fig_simstudy}
\end{center}
\end{figure}

Results are shown in Figure~\ref{Fig_simstudy}, where we show the average (relative) scores and the 5\%-95\% range over the $N_\text{rep}=50$ repeated draws. When the underlying data are non-sparse and stationary (Scenario 1), i.e., M1 is the ``correct'' model, all methods except covariance tapering perform \newTxt{nearly} as well as M1. \st{in terms of both CRPS and RMSE.} 
This makes sense for the Vecchia approximation, since much of the underlying theory is derived for traditional (i.e., non-sparse) stationary kernels; similarly, the predictive process has been shown to perform well in capturing large-scale features (as is present with a non-sparse kernel). For the same reason, it is unsurprising that covariance tapering performs worse than the other methods for non-sparse data, as in Scenario 1. It is reassuring that our kernel (which is nonstationary by definition for both M5 and M6) performs well for stationary data. 
\newTxt{A slight exception to this is when the SNR is low: in this case, M5 and M6 perform worse, on average, than M1. This too is unsurprising, because when the noise is large and the ground truth is smooth, a sparse kernel could overfit the noise. This has important implications for the sorts of data sets that may most benefit from our proposed kernel: since sparsity is a key feature of the kernel, it may not be best suited to very smooth functions.}

\st{Interestingly} \newTxt{For Scenario 2, where the ground truth is sparse and stationary, M2 and M4 perform nearly as well as M1, while M3 performs significantly worse (indicating that the predictive process struggles to capture sparse structure, as has been observed elsewhere). M5 and M6 perform as well as M1 when the SNR is large, but start to show large increases in performance (on average, 10-15\%) when the SNR is small. }
\st{, M5 and M6 perform similarly--albeit slightly better--than M1 when the underlying data are sparse and stationary (Scenario 2), while the Vecchia approximation (M2) and covariance tapering (M4) perform slightly worse, on average, than an exact stationary GP (M1). For sparse data, it is clear that the predictive process cannot capture the finer-scale structure.} 
However, when the underlying true state involves nonstationarity (Scenarios 3 and 4), the nonstationary kernels (M5 and M6) yield a noticeable improvement over a stationary GP, \st{particularly in terms of CRPS} \newTxt{particularly for low SNR}. In contrast, M2 and M3 are noticeably worse for these scenarios, which suggests that the standard configuration of the sparse general Vecchia approximation and the predictive process is sub-optimal for non-stationary data. The advantage of our kernel is particularly noticeable when the underlying ground truth is nonstationary \textit{and} sparse.

In summary, these comparisons verify that our kernel performs as well or better than more traditional kernels within a GP across a variety of ground truth data-generating mechanisms and a range of SNR values. 
\newTxt{The improvements in performance are admittedly modest for SNR = 10 or 20, but for SNR = 5 the relative improvement for M5 and M6 vs. M1 is on 12-15\% on average and as large as 50\% for some synthetic data sets. In other words, for large noise, when the underlying data are sparse and/or nonstationary, the stationary Mat\'{e}rn fails to characterize the true function; however, even in these cases, M5 and M6 can capture the ground truth.}
This furthermore verifies that each of the data-generating mechanisms (different combinations of stationary/nonstationary and sparse/nonsparse) are special cases of our kernel (as described in Section~\ref{subsec:specialcases}); in other words, our kernel can learn the true underlying state of a given data set.

\section{Application: space-time prediction of daily \newTxt{maximum} temperature} \label{sec:application}

Probabilistic machine learning methods like GPs are well-suited for conducting spatio-temporal prediction of in situ measurements of daily maximum temperature. While these data are the gold standard in the Earth sciences, their irregular sampling presents a challenge for climate and atmospheric scientists, who often require data on regular space-time units. To address this need, there are a large number of so-called gridded data products that use (in many cases) na\"ive statistical methods to conduct spatio-temporal interpolation. For example, the Livneh et al. (2015) \cite{livneh2015spatially,livneh2015dataset} data product (henceforth L15) provides estimates of \newTxt{daily maximum} surface temperature on a regular grid over the United States from 1950 to present. The L15 data are generated using inverse-distance weighted averaging \citep{shepard1984computer} separately for each day's measurements before applying a post-hoc correction for orography; no measures of uncertainty quantification are provided.
GPs provide many value-added benefits to conducting space-time prediction for \newTxt{daily maximum} temperature measurements relative to the L15 methodology: data-driven (not data-agnostic) weights for spatial averaging; nonstationary weights for interpolation; the ability to account for multiple-day autocorrelation in \newTxt{daily maximum} temperature records; a flexible prior mean function to capture known spatio-temporal structure and systematically estimate relationships between daily \newTxt{maximum} temperature and orography; and most importantly uncertainty quantification. 

\subsection{Spatial modeling of data \newTxt{from individual days}} \label{sec:spatmod}

We first apply the kernel $C_z$ and the Bayesian model to \st{time slices} \newTxt{all measurements from a given day} of daily maximum temperature ($^\circ$C) from the GHCN-D database \citep{Menne2012}, considering stations that lie within the contiguous United States (CONUS). These measurements are the same input data used to construct the L15 data product. We identify a set of $N_\text{space} = 1174$ GHCN-D stations with high-quality records (see Figure~\ref{Fig_priormean}a.) and randomly select 60 days for training: one from each month over 2001-2005. In this case, the input space collapses down to two-dimensional coordinates, longitude and latitude, such that $\mathcal{X} \subset \mathbb{R}^2$.
For test data, we use measurements from $N_\text{space, test}=1444$ records not included in the training records.

We refer the interested reader to Appendix~\ref{apdx:spatmod} for full details on the prior mean function and specific kernel used for \newTxt{spatial modeling} \st{the time slices}. \st{In summary, the prior mean is a linear function of orographic features and distance to the nearest coastlines to account for known relationships between these features and daily temperature.} \newTxt{In summary, in order to maintain a fair comparison with the L15 methodology, the prior mean is set to be a constant.} The core kernel is the locally anisotropic nonstationary kernel defined in \cite{risser2020bayesian} wherein the signal variance and anisotropy length scales are suitably transformed linear functions of elevation \st{and distance to the coast}. 
For the sparsity-inducing kernel, following the guidance of Section~\ref{sec:sde1} we err on the side of choosing $n_1$ and $n_2$ to be larger than we might expect. We consider three configurations, each with 100 total bump functions: (1) $n_1 = 2$ and $n_2=50$; (2) $n_1=4$ and $n_2 = 25$; and (3) $n_1 = 10$ and $n_2 = 10$. Furthermore, we consider three upper limits on the Wendland truncation radii (5, 10, and 15 units of degrees longitude/latitude) to ensure that the resulting kernel has the chance to impose sparsity on the data. Ultimately, our specification results in \st{five} \newTxt{one} hyperparameter for the prior mean function, \st{twelve} \newTxt{eight} hyperparameters for the core kernel, one hyperparameter for the noise variance, and $3 + n_1n_2(d+3) = 3 + 100\times5 = 503$ hyperparameters for the sparse kernel: in total there are $1 + 8 + 1+503 = 513$ hyperparameters for the GP applied to each \newTxt{day's measurements} \st{time slice}.

\begin{figure}[!t]
\begin{center}
\includegraphics[trim={0 0 0 0mm}, clip, width = \textwidth]{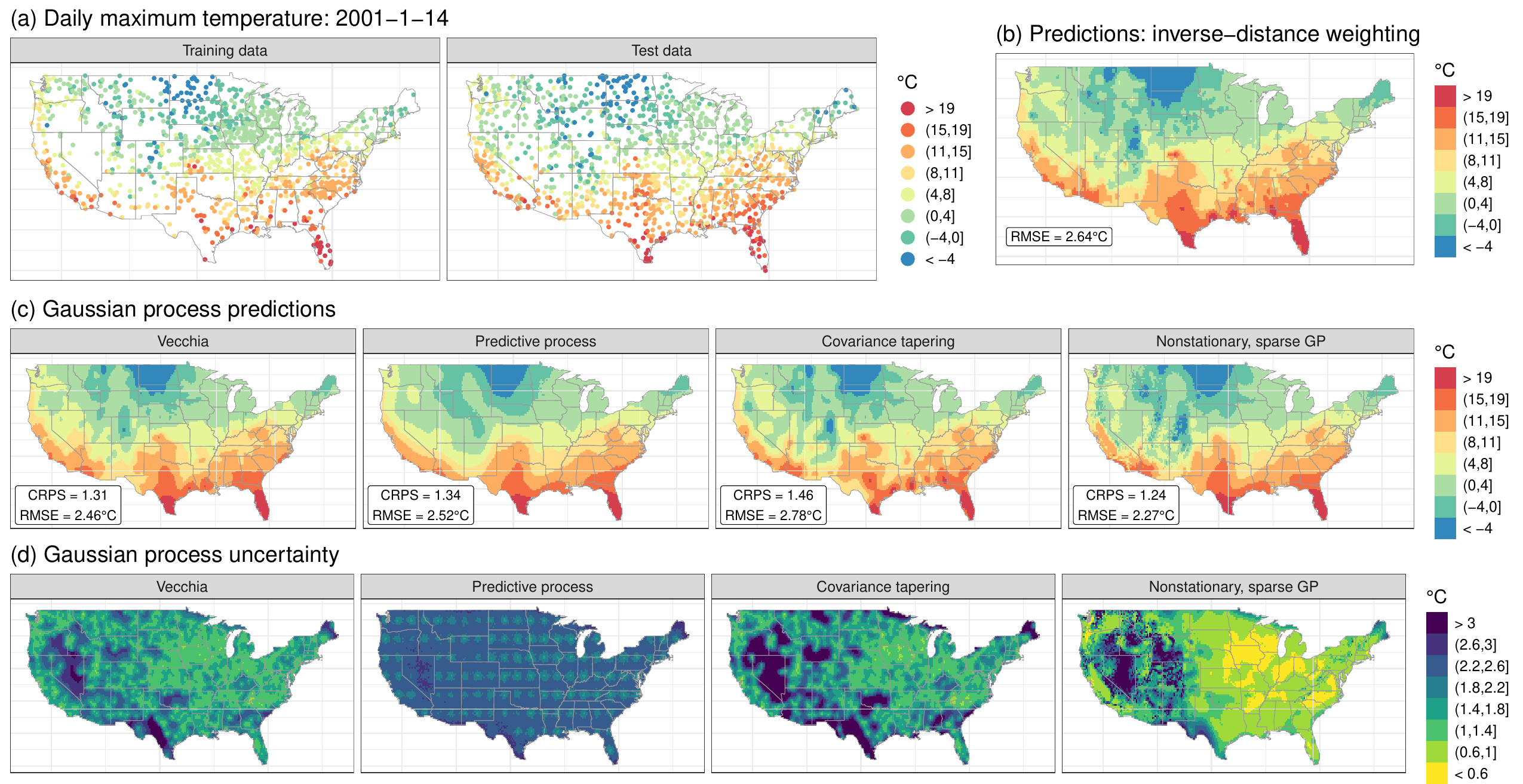}
\caption{\st{Training and test data for the January 14, 2001 time slice of daily maximum temperature (panels a. and b., respectively). Panels (c) and (d) show the spatial distribution of interpolated daily maximum temperature for both the inverse-distance weighting and our method. The inset text in panels (c) and (d) shows the RMSE for the test data using each method.}
\newTxt{Training and test data for January 14, 2001 daily maximum temperature (panel a.). Panels (b) and (c) show the spatial distribution of interpolated daily maximum temperature for inverse-distance weighting and the various Gaussian process methods, while panel (d) shows the Gaussian process uncertainties (posterior standard deviation). The inset text in panels (b) and (c) shows the RMSE and CRPS for the test data using each method.}
}
\label{Fig_timeslice2}
\end{center}
\end{figure}

\vskip1ex
\noindent \textbf{Results.} 
Before analyzing all 60 days \st{time slice}, we first looked at \newTxt{data from} \st{the} January 14, 2001 \st{time slice} (training and test data shown in Figure~\ref{Fig_timeslice2}) to determine an optimal configuration of $n_1$ and $n_2$ as well as the Wendland length scale upper bound. Three $n_1 \times  n_2$ configurations ($10\times 10$, $4\times 25$, $2\times 50$) and three upper bound thresholds result in nine different sparse-inducing GP models. For each model, the MCMC was run for 20,000 iterations, with the first 15,001 discarded as burn-in; convergence of the MCMC was assessed graphically with no issues detected \newTxt{(trace plots from the $4\times 25$ configuration are shown in Figures~\ref{Fig_trace1} and \ref{Fig_trace2})}. All subsequent calculations were based on the posterior median of the hyperparameters. The log of the marginal likelihood of the training data, conditioned on the posterior median, the resulting sparsity of the training data covariance, as well as out-of-sample root mean square error (RMSE) and the continuous rank probability score \citep[CRPS;][]{Gneiting2007} for the test data are shown in Supplemental Figure~\ref{Fig_timeslice0} for each of the nine models. For comparison, Supplemental Figure~\ref{Fig_timeslice0} also shows corresponding quantities from a similar Gaussian process model with the same prior mean and core kernel but without the sparse kernel (dashed lines). A few interesting results emerge from the different models considered: first, as expected, the sparsity increases monotonically with the Wendland upper bound; however, unexpectedly, there is a clear trend in the sparsity across the $n_1 \times  n_2$  configurations, wherein larger values of $n_1$ lead to a smaller fraction of nonzero elements in the covariance. The in-sample log-likelihood and out-of-sample RMSE/CRPS all greatly prefer longer Wendland length-scales (upper bound of 10 or 15), and from a testing standpoint the sparse nonstationary GPs perform as well as the nonsparse nonstationary GP (with about $1/3$ the sparsity). As a compromise between good out-of-sample performance and sparsity, we conclude that the $n_1=4$ and $n_2 = 25$ configuration with upper bound 10 is the ``optimal'' configuration to use for in situ measurements of daily maximum temperature.

Using the optimal configuration, we show visualizations of the \st{prior mean and} covariance hyperparameters for the \newTxt{data from} January 14, 2001 \st{time slice} in Figure~\ref{Fig_timeslice1}.  Since the \st{prior mean function and} core kernel parameters are specified as functions of physical, topographical covariates, we can visualize the implied surfaces for the \st{mean function,} core kernel signal variance, core kernel anisotropy length-scales, and implied correlation between a set of reference locations and all other points in a $0.25^\circ \times 0.25^\circ$ longitude-latitude grid over the United States. These quantities (see Figure~\ref{Fig_timeslice1}) provide useful scientific information regarding the \st{first- and} second-order properties of daily \newTxt{maximum} temperatures, and how these vary according to topography. We can furthermore see how the bump functions impose regions of distance-unrelated correlations across the United States. \newTxt{We note that small changes to the reference location can result in very different correlation structures across the rest of the domain. Since our proposed kernel is highly flexible and fully data-driven (with only a relatively small penalty for regularization), it is difficult to say whether these correlations are a property of the underlying data generation mechanism or an artifact related to this specific data sample.}

Finally, we compare predictions for a $0.25^\circ \times0.25^\circ$ longitude-latitude grid in Figure~\ref{Fig_timeslice2} for the L15 inverse-distance methodology with corresponding GP-based estimates; see Figure~\ref{Fig_timeslice2}(c) and (d). \newTxt{In addition to our new methodology, we also show GP-based estimates from models M2 (Vecchia), M3 (predictive process), and M4 (covariance tapering), since the L15 methodology is a deterministic, non-model-based technique that does not account for covariates or uncertainty.} Our methodology yields a clear improvement over L15 \newTxt{and the other GP predictions}: both with respect to reduced root mean square error \newTxt{and CRPS} for the test data, and also for interpolation of daily maximum temperature in mountainous terrain. For example, note that the inverse-distance weighting and competing GP methods completely fail to characterize the topographic variability in, e.g., Nevada and California. Furthermore, it is clear that the L15 method ``under-smooths'' the data in the eastern United States, particularly along the Gulf Coast. \newTxt{Our nonstationary, sparse GP also yields heterogeneous uncertainty estimates, reflecting the underlying differences in the degree of uncertainty in temperature measurements in flat versus topographically diverse regions.}

Now stepping back to consider all 60 days \st{time slices}, we show \st{corresponding in-sample log-likelihood, sparsity, and} out-of-sample prediction metrics for our proposed method in Table~\ref{tab_all_timeslices_new}. For reference, we also summarize RMSE for inverse distance weighting used by the L15 method \newTxt{and the other GP-based methods}. It is once again clear that a sparse nonstationary GP vastly outperforms L15: on average, the errors are roughly $100\%(3.00-2.62)/3.00 = 12.6\%$ smaller. \st{(furthermore, the nonstationary GP errors are uniformly smaller than the L15 errors across all days time slices). As before, the $n_1=4$ and $n_2=25$ configuration results in about 37\% sparsity, on average.} \newTxt{Furthermore, our nonstationary GP outperforms the other GP-based methods considered here, providing the best RMSE and CRPS on average.}


\begin{table}[!t]
\begin{tabular}{l|rrr|rrr}
 & \multicolumn{3}{c}{\textbf{RMSE (test data; $^\circ$C)}} & \multicolumn{3}{c}{\textbf{CRPS (test data; no units)}} \\
\textbf{Method} & \textit{Average} & \textit{Minimum} & \textit{Maximum} & \textit{Average} & \textit{Minimum} & \textit{Maximum} \\ \hline
Inverse distance weighting &  3.00 & 2.38 & 4.22 & -- & -- & -- \\ 
Vecchia approximation & 2.79 & 2.25 & 3.80 & 1.53 & 1.22 & 2.05\\
Predictive process & 2.84 & 2.30 & 3.80 & 1.54 & 1.23 & 2.07\\
Covariance tapering & 3.04 & 2.33 & 4.20 & 1.65 & 1.27 & 2.22\\
Nonstationary, sparse GP &  2.62 & 2.02 & 3.74 & 1.50 & 1.09 & 2.15 \\
\end{tabular}
\vskip1ex
\caption{Out-of-sample root mean square error (RMSE) and continuous rank probability score (CRPS) for inverse-distance weighting, three competing methods, and our new methodology. Metrics are aggregated over all 60 days, showing the average, minimum, and maximum taken over the 60 days. Note: CRPS cannot be calculated for inverse distance weighting because the methodology does not yield uncertainty measures.
}
\label{tab_all_timeslices_new}
\end{table}

\subsection{Spatio-temporal modeling} \label{sec:spattimemod}

We next apply the Bayesian model proposed in Eq.~\ref{CANONmodel} to spatio-temporal measurements from the $N_\text{train}=1174$ stations over 2001-2005 ($N_\text{time}=1826$ days), such that the input space is $\mathcal{X} = \mathcal{S} \times \mathcal{T}$ (two-dimensional space and one dimension for time), i.e., $d=3$. To make the calculation more feasible within the allotted time frame, we pick every second data point which yields a total of $N=1,038,723$ non-missing daily measurements for training.
We again refer the interested reader to Appendix~\ref{apdx:spattimemod} for full details on the prior mean function and specific kernel used for the spatio-temporal measurements. In summary, in light of the fact that we are modeling high-dimensional space-time data and our kernel can discover and capitalize upon sparsity, it is to our benefit to propose a prior mean function that explains as much of the known spatial and temporal structure in the data as possible. The most prominent structure in a data set involving temperature is the annual cycle, wherein daily maximum temperatures in the Northern Hemisphere are generally lowest in the winter and highest in the summer. As in Section~\ref{sec:spatmod}, there is an underlying geospatial climatology to measurements of \newTxt{daily maximum} temperature including an implicit zonal structure wherein spatial locations farther from the equator are generally cooler, a strong influence from orographic variability, and the influence of coastlines. 

To account for each of these features, we propose a flexible prior mean function using natural cubic splines with spatially-varying coefficients defined by thin-plate spline basis functions and orography. \st{After some trial and error and i} \newTxt{I}n order to maintain a compromise between sufficient flexibility and overfitting, we end up with $1899$ hyperparameters for the prior mean: one for every $\approx 550$ data points. Using ordinary least squares, the prior mean function explains 94.1\% of the variability in the daily \newTxt{maximum} temperature data and removes meaningful autocorrelations beyond $\approx 5$ days (see Figure~\ref{Fig_priormean}). In other words, the variability explained by a flexible prior mean function ``increases'' the sparsity that can be leveraged by the kernel. 
The space-time core kernel is similar to the one in Equation~\ref{eq:cov_timeslice}, albeit generalized to a three-dimensional input space, i.e., $C_\text{core}(\bfx, \bfx')$, with stationarity in the time direction. It is important to note that this specification yields a non-separable space-time kernel, i.e., $C_\text{core}(\bfx, \bfx') \neq C(\bfs, \bfs') \times C(t, t')$.
For the sparse kernel, the Wendland component is again anisotropic with coordinate-specific length scales (and coordinate-specific truncation limits); following Section~\ref{sec:spatmod} we use 10 units for the spatial truncation limits and 4 days for the temporal truncation limits based on the fact that the prior mean function explains autocorrelation over longer time lags. We again use 100 bump functions with $n_1=4$ and $n_2=25$; however, based on the fact that the prior mean explains so much of the temporal variability, we restrict the bump functions to live in the spatial domain only. This yields $521 + 1 + 1= 523$ hyperparameters for the kernel; combined with the prior mean hyperparameters, we have $2,422$ hyperparameters to be learned in training with MCMC.

\vskip1ex
\noindent \textbf{Results.}
Training the final GP involved 1000 MCMC iterations, which translated into approximately 7000 likelihood evaluations on 256 NERSC Perlmutter A100 GPUs. Although the time cost varies slightly based on the current hyperparameters and the induced sparsity structure of the covariance matrix, generally speaking, a representative likelihood took 66 seconds to compute: 37 seconds for distributed covariance computation, about 9 seconds for the MINRES (Minimal Residual Method, \citep{paige1975solution}) linear solve, 2 seconds for the log-determinant calculation, with the rest of the time spent on mean and noise function calculations and necessary transformations between formats. The observed sparsity was approximately 0.0006 (nonzero elements/matrix size). In total, this led to an approximate computation time of 128 hours -- well in line with the training of large neural networks or large language models. 

\begin{figure}[!t]
\begin{center}
\includegraphics[trim={0 0 0 0mm}, clip, width = \textwidth]{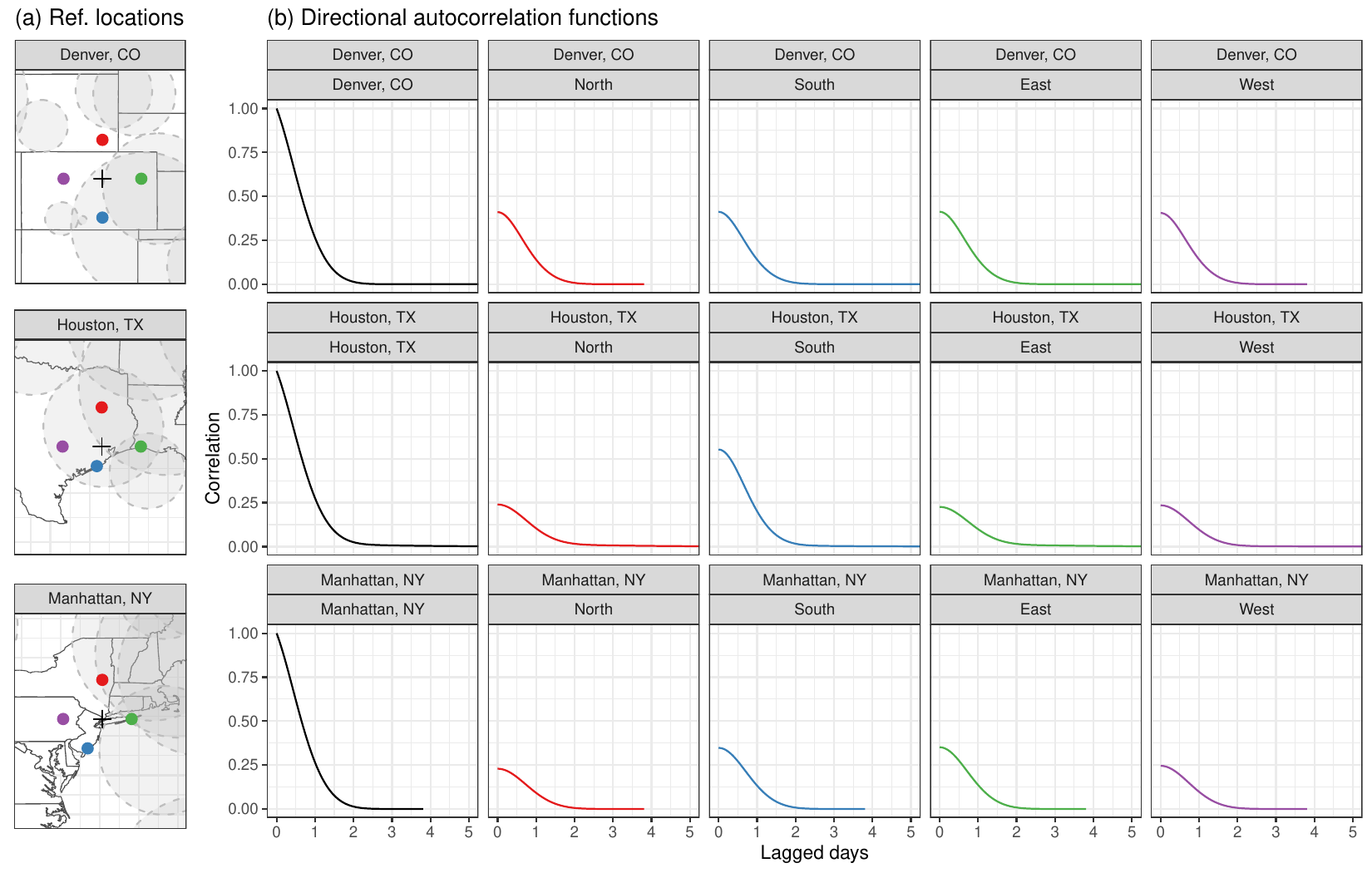}
\caption{Directional autocorrelation functions for pairs of locations over time. Here, we focus on three reference locations: Denver, CO; Houston, TX; and Manhattan, NY. For each of these reference locations, we compare correlation at four sites that are $2^\circ$ north, south, east, and west. In panel (a), the activated bump functions (i.e., those with $a_{ij}=1$) are shown in dashed gray. In panel (b), correlations are only plotted when they are nonzero. 
}
\label{Fig_corr_fcns}
\end{center}
\end{figure}

\begin{figure}[!t]
\begin{center}
\includegraphics[trim={0 0 0 0mm}, clip, width = \textwidth]{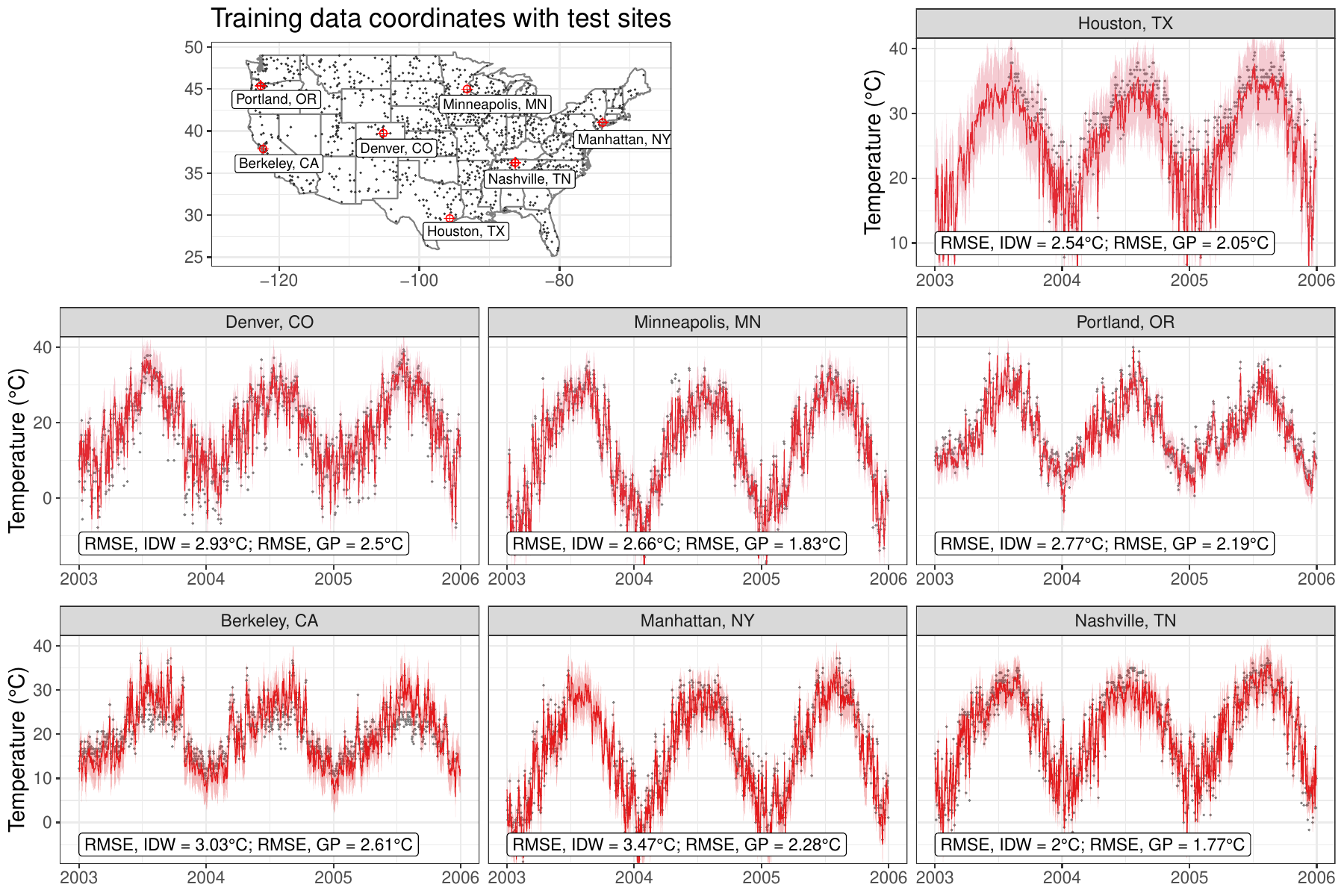}
\caption{Test data with posterior predictions and uncertainties for seven selected selected sites over 2003-2005 (2001-2002 omitted for visibility of results). The shaded band represents the three-sigma uncertainty interval.
}
\label{Fig_spacetime}
\end{center}
\end{figure}

To summarize the results of this analysis, we selected seven test sites from across the United States \newTxt{(see the top left panel of Figure~\ref{Fig_spacetime}).
We first explore the spatio-temporal correlations implied by our sparse, nonstationary kernel; for brevity, we focus on three of these sites and plot directional autocorrelation functions in Figure~\ref{Fig_corr_fcns} (corresponding plots for the four other sites are shown in Figure~\ref{Fig_corr_fcns2}). Here, we define directional autocorrelation functions to be the time-evolving correlation for a pair of sites, i.e., $C_y\left( [\bfs, t], [\bfs', t']  \right)$ (normalized to show correlation instead of covariance), where $\bfs$ is one of the reference cities and $\bfs'$ is a point to the north, south, east, or west of the reference; since our kernel is stationary in the time domain we set $t = 0$ and $t' \in [0, 5]$. Figure~\ref{Fig_corr_fcns}(a) also shows the activated bump functions (i.e., those with $a_{ij} = 1$) and their size; recall that the bump functions are only defined in the spatial domain. The three selected sites each have unique correlation properties. The site of interest at Denver, CO is located within an activated bump function, and hence its autocorrelation function decays to zero but is never exactly zero. The same is true for the autocorrelation with points to the south and east, since these points share at least one bump function with Denver, CO. However, points to the north and west do not share a bump function, which means that their autocorrelation goes to exactly zero at the Wendland truncation radius in the time dimension (the posterior median of which is 3.8 days). Importantly, this demonstrates that defining bump functions only in the spatial domain (and not time) in combination with a dense core kernel (the Mat\'{e}rn) results in local non-sparsity. The reference site at Houston, TX is located inside an activated bump function and shares at least one bump function with the four comparison points; therefore, their autocorrelation is again never identically zero. Finally, the reference site near Manhattan, NY is not contained within an activated bump function and hence shares a bump function with none of the four comparison sites. This means that the autocorrelation between all pairs of sites is exactly zero after 3.8 days. Similar behavior emerges for the other four sites of interest.
}

\newTxt{Next, for the same test sites, we} assess the time-evolving behavior of daily maximum temperatures over 2001-2005; see Figure~\ref{Fig_spacetime} (the first two years are omitted for visibility of results). 
\newTxt{For this very large data set, taking draws from the posterior predictive distribution described in Section \ref{subsec:postPred} is computationally infeasible; instead, we simply calculate the kriging mean and covariance ${\bf m}_{{\bf y}|{\bf z}}$ and ${\bf C}_{{\bf y}|{\bf z}}$  using the posterior median of $\bfbeta$ and $\bftheta$.}
The training sites with test locations are shown on the top left, with the test data and posterior predictions shown in each thumbnail. The most obvious feature of the predictions is the annual cycle, which is well-characterized by our prior mean function. However, note that the specific shape of the annual cycle is flexibly modeled across sites: \newTxt{daily maximum} temperatures in Houston, TX (top right) increase sharply in late winter and decrease sharply in late fall, while temperatures in Portland, OR (middle left) increase and decrease much more slowly between the cold and warm season. The GP predictions maintain the large day-to-day variations present in daily maximum temperatures, which is reflected in the data. Also note that the borrowing of strength over space and time allows the GP to learn the time-varying structure even when there are large chunks of missing data in the training data, e.g., in the first half of 2003 at the Houston, TX gauged location. 
\newTxt{Our spatio-temporal modeling provides a significant benefit in terms of prediction accuracy relative to inverse-distance weighting: as shown in each thumbnail of Figure~\ref{Fig_spacetime}, the improvement in RMSE ranges from 11.5\% (at the site near Nashville, TN) to more than 30\% (at the site near Manhattan, NY site).}
Finally, the three-sigma GP uncertainties, represented by the red shaded band in Figure~\ref{Fig_spacetime}, are clearly nonstationary as well: note the larger magnitude of uncertainties at, e.g., Houston, TX compared to the smaller uncertainties for the Berkeley, CA site.

\section{Discussion} \label{sec:discussion}

In this work, we have proposed a highly flexible kernel that encodes and discovers both sparsity and nonstationarity. We demonstrate how the proposed kernel can be embedded within a Bayesian Gaussian process stochastic model, as well as flexible MCMC tools for GP training. Our methodology outperforms a variety of competing methods across different ground truth data and furthermore provides better prediction errors than state-of-the-art methods in the Earth sciences for measurements of daily maximum temperature. Our application of an exact GP to one million data points is on par with the largest GPs applied to date. In the scope of exact GPs, \cite{wang2019exact} scaled to 1.3 million data points by taking advantage of GPU-accelerated matrix-vector products in Conjugate-Gradient solver for faster training. For this method to work, the covariance matrix has to be well-conditioned which is only the case for some stationary kernels. In addition, their method has a natural maximum scaling due to the sheer size of the dense covariance matrix. Our proposed method, in contrast, takes advantage of naturally occurring sparsity, has shown favorable weak and strong scaling properties \cite{Noack2023}, and works in harmony with non-stationary kernels. 

Depending on the level of sparsity that is discovered by the proposed kernel, computational challenges arise during the log-determinant calculation and the linear system solution. The log-determinant calculation is currently performed using a Lanczos quadrature algorithm \citep{Ubaru2017}. Although this estimation is fast, it is approximate, which may or may not affect the number of MCMC iterations and the solution; however, among methods to train a GP, the MCMC will likely handle this inaccuracy most robustly. Ideally, one might deploy a sparse Cholesky or LU decomposition, but current algorithms produce (LU and Cholesky) factors that are not sparse enough to be stored. We are working on a Python LU multi-node decomposition that would allow fast and exact computation of the log-determinant. LU or Cholesky decomposition would also allow for a fast and exact linear system solution, which is currently accomplished by the MINRES algorithm. Given our kernel and covariance structure, the conjugate-gradients method converges poorly, and common preconditioners are more time-consuming than MINRES, in our case. 
    
With regards to scalability beyond one million data points, \citep{Noack2023} derived a formula to predict how much time it would take to calculate the covariance matrix for $N$ data points: $T=\frac{N^2 t_b}{2nb^2}$.
Here, $N$ is the size of the dataset, $t_b$ is the time it takes for a covariance batch to be computed, $n$ is the number of workers, and $b$ is the batch size. The validity of this formula has been well-tested in our experiments. Assuming $n=20,000$ available GPUs that can calculate a batch of size $b=15,000$ in 2 seconds, we could calculate the covariance matrix for 100 million data points in 2222 seconds; with the current algorithm, the log-determinant and the MINRES could be calculated in about 1000 seconds each. This results in a log marginal likelihood evaluation time of just over an hour. Whether or not this is still in the realm of feasibility depends on the total available computing time. If the number of hyperparameters is relatively small, the MCMC might converge after around 500 iterations, which amounts to just over 3 weeks of total computing time. This toy example reveals that a 100 million-data-point GP is on the horizon. Note, however, that this calculation assumes constant sparsity similar to the one observed in the real data application in Section \ref{sec:spattimemod}. Early evidence suggests that larger datasets will allow the proposed kernel to discover more sparsity. While this will not affect the calculation of the covariance matrix, it vastly reduces linear system solutions and log-determinant calculations.

\backmatter

\section*{Declarations}

\noindent \textit{Funding.} This research was supported by three programs:
\begin{enumerate}
    \item The Director, Office of Science, Office of Biological and Environmental Research of the U.S. Department of Energy under Contract No. DE-AC02-05CH11231 and by the Regional and Global Model Analysis Program, and the Calibrated and Systematic Characterization, Attribution, and Detection of Extremes (CASCADE) Scientific Focus Area, within the Earth and Environmental Systems Modeling Program.
    \item The Center for Advanced Mathematics for Energy Research Applications (CAMERA), which is jointly funded by the Advanced Scientific Computing Research (ASCR) and Basic Energy Sciences (BES) within the Department of Energy’s Office of Science, under Contract No. DE-AC02-05CH11231. 
    \item This research was supported in part by the U.S. Department of Energy, Office of Science, Office of Advanced Scientific Computing Research's Applied Mathematics program under Contract No. DE-AC02-05CH11231 at Lawrence Berkeley National Laboratory.
\end{enumerate} 
The research also used resources of the National Energy Research Scientific Computing Center (NERSC), also supported by the Office of Science of the U.S. Department of Energy, under Contract No. DE-AC02-05CH11231.

\vskip1ex \noindent \textit{Conflict of interest/Competing interests.} The authors declare no competing financial interests.

\vskip1ex \noindent \textit{Ethics approval  and consent to participate.} Not applicable.
\vskip1ex \noindent \textit{Consent to participate.} Not applicable.
\vskip1ex \noindent \textit{Consent for publication.} Not applicable.

\vskip1ex \noindent \textit{Availability of data and materials.} The in situ \newTxt{daily maximum} temperature records supporting this article are based on publicly available measurements from the National Centers for Environmental Information (\url{ftp://ftp.ncdc.noaa.gov/pub/data/ghcn/daily/}).

\vskip1ex \noindent \textit{Code availability.} All data analysis in this manuscript was conducted using open-source programming languages and software (namely, R and Python). The \pkg{BayesNSGP} software package for \proglang{R} is available at \url{https://github.com/danielturek/BayesNSGP}. The \pkg{gpCAM} Python package is available  via ``pip install gpcam'' or at \url{https://github.com/lbl-camera/gpCAM}.

\vskip1ex \noindent \textit{Authors' contributions.} 
MDR contributed initial conception of the study and led all analyses, figure generation, and writing.
MMN contributed to the initial conception, developed the Python algorithm for scalable GPs, ran all models on the Perlmutter supercomputer, and added various content to the manuscript.
HL contributed to discussion of the methodological development and improvement of mixing, and writing. 
RP contributed by optimizing the performance of the scalable-GP code and streamlined the code deployment on Github and PyPI. 
All authors contributed to the interpretation of study results and editing of the study's writing and figures.

\bibliography{sn-bibliography} 

\clearpage

\begin{appendix}
\numberwithin{equation}{section}
\numberwithin{figure}{section}
\numberwithin{table}{section}

\section{Proof of Propositions} \label{sec:proof}

\noindent \textbf{Proposition 1.} \textit{The kernel $C_\text{sparse}$ is strictly positive definite. Furthermore, $C_y$ is strictly positive definite whenever $C_\text{core}$ is. More formally, for every admissible parameter vector} 
\[
\vartheta=\bigl(s_{0},r_{0},\{h_{ij},a_{ij},b_{ij},r_{ij}\}\bigr)\quad\text{with }s_{0}>0,
\]
\textit {the kernel} 
\begin{equation}
C_{\text{sparse}}(x,x';\vartheta)=s_{0}\,f_{0}\!\bigl(x,x';r_{0}\bigr)+\sum_{i=1}^{n_{1}}\Bigl[\sum_{j=1}^{n_{2}}g_{ij}(x;\eta_{ij})\Bigr]\,\Bigl[\sum_{j=1}^{n_{2}}g_{ij}(x';\eta_{ij})\Bigr]\label{eq:C_sparse2}
\end{equation}
\textit{is \emph{strictly} positive-definite (PD) on every finite subset of
$\mathcal{X}\subset\mathbb{R}^{d}$. }
\begin{proof}
For each fixed $i$ define 
    \[
    \phi_{i}(x)=\sum_{j=1}^{n_{2}}g_{ij}\bigl(x;\eta_{ij}\bigr),\qquad K_{i}(x,x')=\phi_{i}(x)\,\phi_{i}(x').
    \]
    Given points $x_{1},\dots,x_{m}$, the Gram matrix $[K_{i}(x_{p},x_{q})]_{pq}$
    equals $\Phi_{i}\Phi_{i}^{\top}$ with $\Phi_{i}=(\phi_{i}(x_{1}),\dots,\phi_{i}(x_{m}))^{\top}$
    and is therefore PSD; see  Section 2.2 of \cite{williams2006gaussian}. 
    Because the sum of PSD kernels is PSD \citep[Proposition 2.3.3 of][]{van2012harmonic}, 
    \[
    K_{\text{bump}}(x,x')=\sum_{i=1}^{n_{1}}K_{i}(x,x')
    \]
    is PSD for \emph{all} choices of $\eta_{ij}=(h_{ij},a_{ij},b_{ij},r_{ij})$.
    Then we come to examine the compactly supported function $f_{0}(x,x';r_{0})=\varphi\!\bigl(\|x-x'\|/r_{0}\bigr)$
    with $\varphi$ as in \citep[Theorem 4.9][]{Wendland1995}, 
    which is strictly
    PD on $\mathbb{R}^{d}$. It is clear that multiplying by $s_{0}>0$
    preserves strictness.
    
    If $K_{1}$ is strictly PD and $K_{2}$ is PSD, then $K_{1}+K_{2}$
    is strictly PD because $v^{\top}(K_{1}+K_{2})v\ge v^{\top}K_{1}v>0$
    for any non-zero vector $v$. Applying this to $s_{0}f_{0}$ and $K_{\text{bump}}$
    shows that $C_{\text{sparse}}$ in \eqref{eq:C_sparse2} is strictly
    PD. So far we require only $s_{0},r_{0},r_{ij}>0$; no further constraints
    are placed on $h_{ij},a_{ij},b_{ij}$. Consequently, every parameter
    draw encountered during optimization or MCMC, and every prediction
    location, retains strict positive-definiteness. The model uses the
    Hadamard product $C_{y}(x,x')=C_{\text{core}}(x,x';\theta_{\text{core}})\,C_{\text{sparse}}(x,x';\vartheta)$.
    By the Schur product theorem \citep[Theorem 5.2.1 of][]{horn2012matrix} 
    the element-wise product of positive- definite matrices is positive-definite,
    so $C_{y}$ is strictly PD whenever $C_{\text{core}}$ is.
\end{proof}
\color{black}

\noindent \textbf{Proposition 2.} \textit{The kernel $C_y$ has $w = \min\{w_0, w_\text{core}\}$ continuous derivatives at zero, where $f_0$ has $w_0$ continuous derivatives at the origin and $C_\text{core}$ has $w_\text{core}$ continuous derivatives at zero.}
\begin{proof}
The proof follows directly from the general Leibniz rule, which generalizes the product rule for higher-order differentials. It is well-established and proven by induction.
The general Leibniz rule states
\begin{equation*}
    \partial^l (gh) = \sum_k^l  \binom{l}{k} (\partial^k g) (\partial^{l-k} h)
\end{equation*}
for two functions $g$ and $h$. If we assume the $g$ and $h$ are $n$ and $m$ times differentiable respectively, the sum only exists for $l \leq \min \{m,n\}$. This can be shown by considering three scenarios:
\begin{align}
    & l \leq \min \{m,n\} \label{proof_scenario1} \\
    & l > \min \{m,n\} \land l < \max \{m,n\} \label{proof_scenario2} \\
    & l \geq \max \{m,n\} \label{proof_scenario3}.
\end{align}
Scenario \eqref{proof_scenario2} and \eqref{proof_scenario3} both lead to terms $\partial^k g$ with $k>n$ or $\partial^{l-k} h$ with $l-k > m$. All terms exist and are continuous only for scenario \eqref{proof_scenario1}, which concludes the proof.
\end{proof}

\subsection{Fixed-domain asymptotics for the bump-product kernel}

Notation follows the manuscript exactly: $\mathcal{X}\subset\mathbb{R}^{d}$
is bounded and convex; the infill design places points $\mathcal{D}_{N}=\{x_{1},\dots,x_{N}\}\subset\mathcal{X}$
with minimum spacing $\delta_{N}=\min_{i\neq j}\|x_{i}-x_{j}\|\to0$
as $N\to\infty$. The data-generating Gaussian process is 
\[
y(x)\sim\mathrm{GP}\!\bigl(0,\;C_{y}^{\star}(x,x')\bigr),\qquad C_{y}^{\star}=C_{\text{core}}(x,x';\theta_{\text{core}}^{\star})\,C_{\text{sparse}}(x,x';\theta_{\text{sparse}}^{\star}),
\]
where $C_{\text{sparse}}$ has the bump-product form of Eq.~(3) in
the paper. All hyper-parameters $\theta_{y}=(\theta_{\text{core}},\theta_{\text{sparse}})$
receive priors with positive, continuous density in a neighbourhood
of $\theta_{y}^{\star}$.
\begin{proposition}
\label{thm:consistency} For every $\varepsilon>0$, assume that the
prior of $\theta_{y}$ have probability mass to every neighborhood
of $\theta_{y}^{\star}$ then its posterior 
\[
\Pi_{N}\bigl(\|\theta_{y}-\theta_{y}^{\star}\|>\varepsilon\bigr)\;\xrightarrow[N\to\infty]{P_{\theta_{y}^{\star}}}\;0,
\]
where $\Pi_{N}$ is the posterior of $\theta_{y}$ given $\bigl(y(x_{1}),\dots,y(x_{N})\bigr)$. 
\end{proposition}

Theorem~2.1 of \cite{van2008rates} gives posterior
contraction whenever (i)~the likelihood is continuous in the finite-dimensional
parameter, (ii)~the prior assigns positive mass to every neighbourhood
of the truth, and (iii)~the model is identifiable locally. Each covariance
entry is a smooth function of $\theta_{y}$, so the likelihood is
continuous; the prior requirement is built in; Proposition~2 of our
current paper guarantees local identifiability. Although the bump
radii $r_{ij}$ constitute a varying support, they lie in a fixed
compact set and the prior places mass near $r_{ij}^{\star}$. Hence
Theorem~2.1 applies verbatim.

\begin{proposition}
[stabilisation of the sparsity pattern] \label{cor:sparsity} Let
\[
r_{\max}^{\star}=\max_{0\le i\le n_{1},\,1\le j\le n_{2}}r_{ij}^{\star},\qquad\delta_{N}=\min_{i\neq j}\Vert x_{i}-x_{j}\Vert,\qquad\Omega_{N}(\theta_{y})=\bigl[C_{y}(x_{i},x_{j};\theta_{y})\bigr]_{i,j=1}^{N}.
\]

For any deterministic sequence $\varepsilon_{N}\downarrow0$ define
\[
K_{N}(\varepsilon_{N})\;=\;\frac{\text{Vol}\!\bigl(B_{d}(2(r_{\max}^{\star}+\varepsilon_{N})+\tfrac{\delta_{N}}{2})\bigr)}{\text{Vol}\mathcal{}\!\bigl(B_{d}(\tfrac{\delta_{N}}{2})\bigr)}\;=\;\Bigl(\tfrac{4(r_{\max}^{\star}+\varepsilon_{N})}{\delta_{N}}+1\Bigr)^{d}.
\]

Let 
\[
\mathcal{A}_{N}=\Bigl\{\theta_{y}:\,\max_{i,j}\vert r_{ij}-r_{ij}^{\star}\vert\le\varepsilon_{N}\Bigr\}.
\]

Then 
\[
\Pi_{N}\!\Bigl(\theta_{y}\in\mathcal{A}_{N}\;\text{and every row of }\Omega_{N}(\theta_{y})\text{ contains }\le K_{N}(\varepsilon_{N})\text{ non-zeros}\Bigr)\;\xrightarrow[N\to\infty]{\;P_{\theta_{y}^{\star}\;}}\;1.
\]
where $\Omega_{N}(\theta_{y})=\bigl[C_{y}(x_{i},x_{j};\theta_{y})\bigr]_{i,j=1}^{N}$. 
\end{proposition}

\textit{Proof.} Proposition~\ref{thm:consistency} implies $\max_{i,j}|r_{ij}-r_{ij}^{\star}|=o_{P}(1)$.
Hence, with high posterior probability, the support of $C_{\text{sparse}}$
is contained in balls of radius $r_{\max}^{\star}+M_{N}\delta_{N}$.
A volume-packing argument bounds the number of design points per ball,
giving the stated uniform row sparsity.

Theorem~1 gives posterior consistency of every bump radius, so $\Pi_{N}(\mathcal{A}_{N})\to1$.
We fix $\theta_{y}\in\mathcal{A}_{N}$ and set $R_{N}=2(r_{\max}^{\star}+\varepsilon_{N})$.
If $C_{\text{sparse}}(x_{i},x_{j};\theta_{y})\neq0$, then either
\[
f_{0}(x_{i},x_{j};r_{0})\neq0\quad\text{or}\quad g_{ij}(x_{i};\cdot)\,g_{ij}(x_{j};\cdot)\neq0.
\]
In both cases $\Vert x_{i}-x_{j}\Vert\le R_{N}$, because $r_{0}$
and every $r_{ij}$ lie in $[0,r_{\max}^{\star}+\varepsilon_{N}]$.
Next we can give a packing argument, and we fix $i$ and set 
\[
\mathcal{N}_{i}(\theta_{y})=\bigl\{ j:\,C_{\text{sparse}}(x_{i},x_{j};\theta_{y})\neq0\bigr\}\subset B_{d}(x_{i},R_{N}).
\]
Because the design is $\delta_{N}$-separated, the closed balls $B_{d}(x_{j},\delta_{N}/2)$,
$j\in\mathcal{N}_{i}(\theta_{y})$, are pairwise disjoint and all
lie inside $B_{d}(x_{i},R_{N}+\delta_{N}/2)$. Hence 
\[
\#\mathcal{N}_{i}(\theta_{y})\;\text{Vol}\bigl(B_{d}(\tfrac{\delta_{N}}{2})\bigr)\;\le\;\text{Vol}\bigl(B_{d}(R_{N}+\tfrac{\delta_{N}}{2})\bigr)\quad\Longrightarrow\quad\#\mathcal{N}_{i}(\theta_{y})\le K_{N}(\varepsilon_{N}).
\]

\noindent On $\mathcal{A}_{N}$ every row of $\Omega_{N}(\theta_{y})$
has at most $K_{N}(\varepsilon_{N})$ non-zeros; since $\Pi_{N}(\mathcal{A}_{N})\to1$,
the corollary follows.

The classical fixed-support proofs (van der Vaart \& van Zanten, Furrer
\emph{et al.}, Stein) require only that the taper radius be finite
and non-shrinking. Proposition~\ref{cor:sparsity} shows that the data-driven
bump radii concentrate on $\{r_{ij}^{\star}\}$, so for large $N$
the support radius is stochastically constant. Conditioning on this
high-probability event allows the existing theorems to be invoked
unchanged; the proposed kernel therefore inherits the full suite of
classical fixed-domain guarantees while discovering its own sparsity
pattern from the data.

\color{black}

\section{Parametric nonstationary kernel} \label{apdx:parnonstat}

In the absence of specific knowledge regarding the form of the core kernel, a natural and flexible choice that performs well in practice is the so-called ``parametric'' nonstationary kernel \citep{noack2024unifying} derived in \cite{Paciorek2006} and later extended in \cite{risser2020bayesian}. This ultra-flexible kernel allows all aspects of the kernel to vary according to positions in the input data:
\begin{equation} \label{PScov}
C_\text{core}(\bfx, \bfx'; \bftheta_\text{core}) = \sigma(\bfx) \sigma(\bfx') 
\frac{\big(\Sigma(\bfx)\Sigma(\bfx')\big)^{d/4}}{\left(\frac{\Sigma(\bfx') + \Sigma(\bfx')}{2} \right)^{d/2}} \mathcal{M}_\nu\left(\sqrt{Q(\bfx, \bfx')}\right), \hskip3ex \bfx, \bfx' \in \mathcal{X},
\end{equation}
where
\begin{equation*} \label{Qij}
Q(\bfx, \bfx') = \frac{(\bfx - \bfx')^\top (\bfx - \bfx')}{\frac{1}{2}[\Sigma(\bfx) + \Sigma(\bfx')] } = \frac{||\bfx - \bfx'||^2}{\frac{1}{2}[\Sigma(\bfx) + \Sigma(\bfx')]},
\end{equation*}
and $\mathcal{M}_\nu(\cdot)$ is the Mat\'ern correlation function with smoothness $\nu$. In Eq.~\ref{PScov}, $\sigma(\cdot)$ is the signal standard deviation, and $\Sigma(\cdot)$ controls the length-scale of correlation. This kernel is locally isotropic; note that it is straightforward to generalize Eq.~\ref{PScov} to be locally anisotropic \citep[see, e.g.,][
]{risser2020bayesian}. One can then parameterize $\sigma(\cdot)$ and $\Sigma(\cdot)$ to be log-linear functions of a set of basis functions, e.g.,
\[
\log \sigma(\bfx) = \log \sigma_0 + \sum_{m=1}^{M_\sigma} s_m(\bfx) \phi_\sigma^m, \hskip5ex \log \Sigma(\bfx) = \log \Sigma_0 + \sum_{m=1}^{M_\Sigma} s_m(\bfx) \phi_\Sigma^m,
\]
where $\sigma_0$ and $\Sigma_0$ are the baseline signal standard deviation and length-scale, respectively, $\{s_m(\cdot): m = 1, \dots, M\}$ is a set of generic basis functions (polynomials, splines, etc.) and $\{\phi_{(\cdot)}^m: m = 1, \dots, M_{(\cdot)}\}$ is a set of basis function coefficients. Of course, the basis functions for $\sigma(\cdot)$ and $\Sigma(\cdot)$ could be different as well. Using this framework, the hyperparameters for the core kernel are then $\bftheta_\text{core} = (\sigma_0, \Sigma_0, \{\phi_\sigma^m\}, \{\phi_\Sigma^m\})$. 

However, when using the generic nonstationary core kernel defined in Equation~\ref{PScov} some care is required to regularize the hyperparameters to guard against overfitting. 
Even if $M_\sigma$ and $M_\Sigma$ are not too large, 
if the ``true'' signal standard deviation and length scale are stationary this could result in overfitting in the case of limited data. To this end, we apply a simple regularization prior to the $\{\phi_{(\cdot)}^m\}$:
\[
\phi_{(\cdot)}^m \stackrel{\text{iid}}{\sim} N(0, v_{(\cdot)}) \hskip3ex \text{where} \hskip3ex v_{(\cdot)} \sim U(0,10^5).
\]
This prior distribution is equivalent to applying an $L_2$ penalty on the $\{\phi_{(\cdot)}^m\}$ as is used in ridge regression in a Frequentist setting; in a Bayesian setting, $v_{(\cdot)}$ is inferred from the data (i.e., it is updated within the MCMC). This prior has the effect of regularizing the basis function coefficients: when $v_{(\cdot)}$ is small, the $\phi_{(\cdot)}^m$ are ``shrunk'' towards zero; on the other hand, when $v_{(\cdot)}$ is large the coefficients are essentially unrestricted.

\section{Gaussian process training via Markov chain Monte Carlo} \label{sec:training}

Generally speaking, the posterior distribution (\ref{posteriorZ}) is not available in closed form regardless of prior choice, and so we must resort to Markov chain Monte Carlo (MCMC) methods to conduct inference on $\bfbeta$ and $\bftheta$ \citep{gilks1995markov}. In practice, the specific Markov chain Monte Carlo algorithm used for any given data set will be customized to find a suitable balance between computational time and efficiency (or ``mixing'') of the chain; however, we found that the following algorithm works well. First, adaptive block Metropolis-Hastings random walk samplers \citep{haario2001adaptive,shaby2010exploring} are used on the following groups of hyperparameters:
\begin{enumerate}
    \item The prior mean function coefficients $\bfbeta$ (note that a closed-form Gibbs update is available for $\bfbeta$ under the likelihood and prior choice described above; we found that an adaptive Metropolis update worked just as well and obviated additional computational steps).
    \item The core kernel hyperparameters $\bftheta_\text{core}$, the error variance hyperparameters $\bftheta_z$, and Wendland hyperparameters $s_0$ and $r_0$.
    \item The bump function positions $\{{\bf h}_{ij}\}$ and radii $\{r_{ij}\}$.
\end{enumerate}
A fourth block updates all of the amplitudes $\{a_{ij}\}$ in a single step. Since these are binary variables, the proposal distribution matches the prior, i.e., a proposed $a_{ij}^*$ is drawn from a Bernoulli distribution with success probability $\pi_{ij}^\text{curr}$ (the current value of the prior probabilities). Lastly, we can use closed-form Gibbs updates for the prior probabilities $\{\pi_{ij}\}$ because their full conditional distribution can be derived in closed form. Specifically, the distribution of a single $\pi_{ij}$ conditioned on all other hyperparameters and the data is
\[
p(\pi_{ij} | {\bf z}, \bfbeta, \bftheta_{-\{\pi_{ij}\}})  \propto  p({\bf z} | \bfbeta, \bftheta) p(\bfbeta,\bftheta)  \propto  p(\pi_{ij}) p(a_{ij} | \pi_{ij})
\]
(here, proportionality refers to quantities involving $\pi_{ij}$). Noting that the prior for $\pi_{ij}$ (uniform on the unit interval) can be re-written as a Beta$(1,1)$ distribution, i.e.,
\[
p(\pi_{ij}) = \frac{\Gamma(1+1)}{\Gamma(1)\Gamma(1)} \pi_{ij}^{(1-1)} (1 - \pi_{ij})^{(1-1)}, \hskip5ex 0 < \pi_{ij} < 1,
\] 
and that the Bernoulli distribution $p(a_{ij} | \pi_{ij})$ is
\[
p(a_{ij} | \pi_{ij}) = \pi_{ij}^{a_{ij}} (1-\pi_{ij})^{(1-a_{ij})},
\]
as a function of $\pi_{ij}$ the product becomes 
\[
\begin{array}{rcl}
    p(\pi_{ij}) p(a_{ij} | \pi_{ij}) & \propto & \pi_{ij}^{(1-1)} (1 - \pi_{ij})^{(1-1)} \pi_{ij}^{a_{ij}} (1-\pi_{ij})^{(1-a_{ij})} \\[0.85ex]
     & \propto & \pi_{ij}^{1 - 1 + a_{ij}} (1-\pi_{ij})^{1 - 1 + 1-a_{ij}}     \\[0.85ex]
     & \propto & \pi_{ij}^{(1 + a_{ij}) - 1} (1-\pi_{ij})^{(2 - a_{ij}) - 1}
\end{array}
\]
which is proportional to a Beta$(1+ a_{ij},2 - a_{ij})$ distribution. Hence, the prior probabilities can be updated by sampling from a Beta$(1+ a_{ij}^\text{curr},2 - a_{ij}^\text{curr})$ distribution.

\section{Synthetic data examples: hyperparameters and sample data sets} \label{appdx:sde}

Here, we provide specific details on the data generating mechanisms S1-S4 used in Section~\ref{sec:sde2}, including hyperparameter configurations and sample draws from each parent distribution. The input space is $\mathcal{X}=[0,10]$, i.e., $d=1$ and elements of the input space are scalars. 
\newTxt{All cases include additive Gaussian white noise wth variance equal to 20\%, 10\%, and 5\% of the signal variance.}

\begin{figure}[!t]
\begin{center}
\includegraphics[trim={0 0 0 0mm}, clip, width = \textwidth]{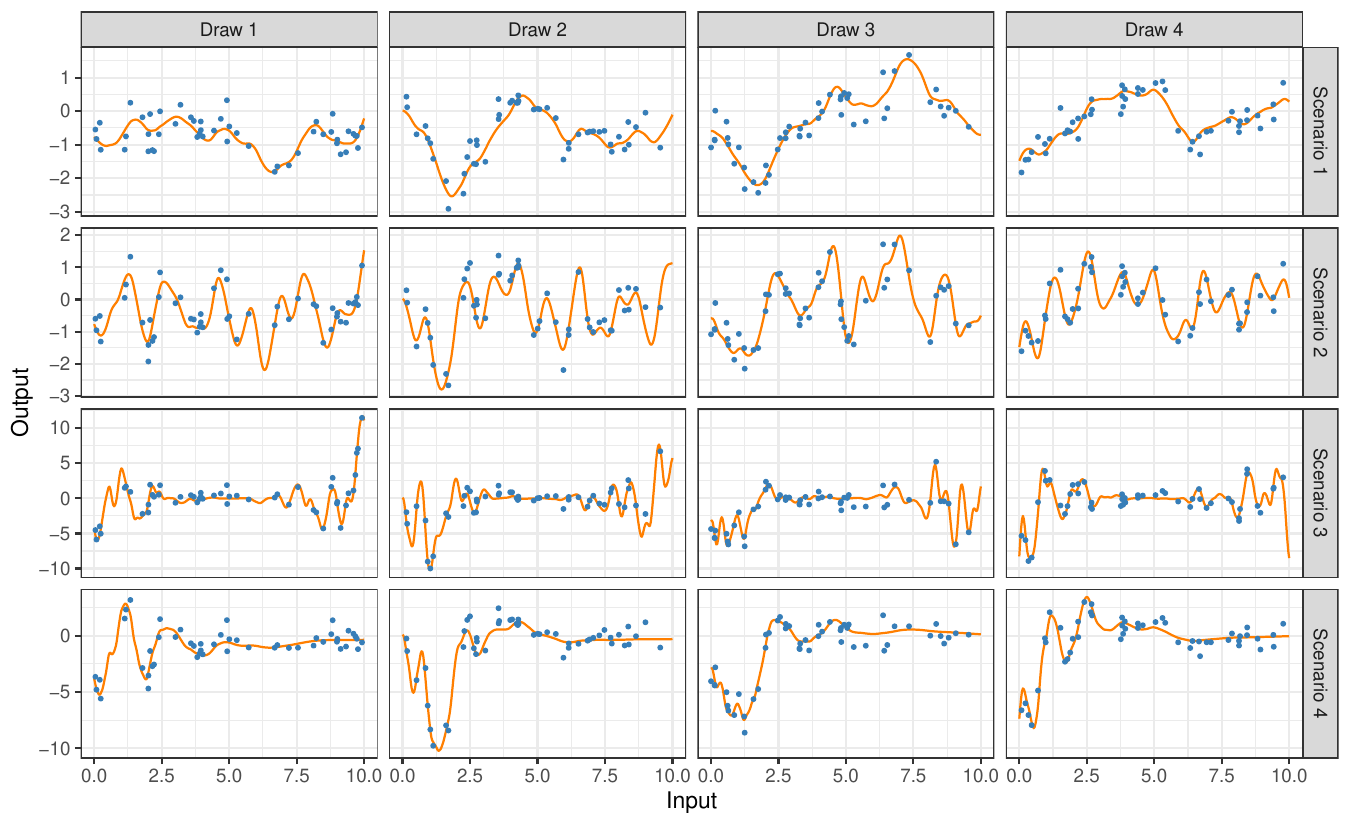}
\caption{Four sample draws from each of the four parent distributions described in Table~\ref{tab:simstudy}.}
\label{SuppFig_sde_examples}
\end{center}
\end{figure}

\paragraph{S1: Non-sparse and stationary.} Here, recall that $C_y(x,x') = \sigma^2 \mathcal{M}_\nu(|x-x'|)$. We specify $\sigma^2 = 1$, $\nu = 2.5$, and the Mat\'ern length-scale parameter is $\rho = 0.5$. 
\st{The training data include additive Gaussian white noise with variance $\tau^2 = 0.1$.}
Sample draws from a mean-zero Gaussian process with corresponding kernel \newTxt{and 10\% noise variance} are shown in Figure~\ref{SuppFig_sde_examples}.

\paragraph{S2: Sparse and stationary.} Here,  $C_y(x, x') = \sigma^2 f_0(x,x'; r_0)$, where $f_0$ is as in Eq.~\ref{eq:compSupp}. We specify $\sigma^2 = 1$ and $r_0 = 1.5$. \st{The training data include additive Gaussian white noise with variance $\tau^2 = 0.1$.} Sample draws from a mean-zero Gaussian process with corresponding kernel \newTxt{and 10\% noise variance} are shown in Figure~\ref{SuppFig_sde_examples}.

\paragraph{S3: Sparse and nonstationary.} Here,  $C_y(x, x') = \sigma(x)\sigma(x') f_0(x,x'; r_0)$, where the signal variance $\sigma^2(\cdot)$ depends on the input locations, $f_0$ is as in Eq.~\ref{eq:compSupp}, and $r_0=0.75$. Specifically,
\[
\sigma^2(x) = 0.05(x-5)^4 + 0.001.
\]
\st{The training data include additive Gaussian white noise with variance $\tau^2 = 0.475$.} Sample draws from a mean-zero Gaussian process with corresponding kernel \newTxt{and 10\% noise variance} are shown in Figure~\ref{SuppFig_sde_examples}.

\paragraph{S4: Non-sparse and nonstationary.} Here, we use
\[
C_y(x, x') = \sigma(x) \sigma(x') 
\frac{\big(\Sigma(x)\Sigma(x')\big)^{1/4}}{\left(\frac{\Sigma(x') + \Sigma(x')}{2} \right)^{1/2}} \mathcal{M}_\nu\left(\sqrt{\frac{|x - x'|^2}{\frac{1}{2}[\Sigma(x) + \Sigma(x')]}}\right),
\]
where $\nu = 2.5$,
\[
\sigma^2(x) = 0.2 \left(\frac{x-10}{3} \right)^4 + 0.1,
\]
and
\[
\Sigma(x) = 0.06\left({x}/{3} \right)^3 + 0.03.
\]
\st{The training data include additive Gaussian white noise with variance $\tau^2 = 0.506$.} Sample draws from a mean-zero Gaussian process with corresponding kernel \newTxt{and 10\% noise variance}  are shown in Figure~\ref{SuppFig_sde_examples}.

\section{Prior mean and kernel for \newTxt{daily maximum} temperature data}

\subsection{Spatial modeling of daily data} \label{apdx:spatmod}


\noindent \textbf{Prior mean function for daily data \st{time slices}.} 
In order to focus on the second-order properties of daily maximum temperature, for the spatial-only modeling we elect to use a spatially-constant prior mean function. We revisit this decision by proposing a more flexible prior mean function for space-time modeling in Section~\ref{apdx:spattimemod}.

\vskip2ex \noindent \textbf{Kernel.} For the core kernel, we use the locally anisotropic nonstationary kernel defined in \cite{risser2020bayesian} wherein the signal variance and anisotropy length scales are suitably transformed linear functions of elevation and distance to the coast. Specifically, for $\bfs, \bfs' \in \mathcal{S}$,
\begin{equation} \label{eq:cov_timeslice}
C_\text{core}(\bfs, \bfs'; \bftheta_\text{core}) = \sigma(\bfs) \sigma(\bfs') \frac{\left|\bfSig(\bfs)\right|^{1/4}\left|\bfSig(\bfs')\right|^{1/4}}{\left|\frac{\bfSig(\bfs') + \bfSig(\bfs')}{2} \right|^{1/2}} \mathcal{M}_{1.5}\left(\sqrt{(\bfs - \bfs')^\top \left(\frac{\bfSig(\bfs) + \bfSig(\bfs')}{2}\right)^{-1}(\bfs - \bfs')}\right), 
\end{equation}
where $\mathcal{M}_{0.5}$ is the Mat\'ern kernel with smoothness equal to 1.5. We parameterize the spatially-varying signal variance as
\[
\log \sigma(\bfs) = \phi_{0} + \phi_{1} \text{Elev}(\bfs) + \phi_{2} \text{Dist2Coast}(\bfs).
\]
For the anisotropy, we use the eigendecomposition
\[
\bfSig(\bfs) = {\bf \Gamma}(\bfs) {\bf \Lambda}(\bfs) {\bf \Gamma}(\bfs)^\top, \hskip3ex \text{where} \hskip3ex {\bf \Lambda}(\bfs) = \left[ \begin{array}{cc} \lambda_1(\bfs) & 0 \\ 0 & \lambda_2(\bfs) \end{array} \right], 
{\bf \Gamma}(\bfs) = \left[ \begin{array}{cc} \cos \gamma(\bfs) & -\sin \gamma(\bfs) \\ \sin \gamma(\bfs) & \cos \gamma(\bfs) \end{array} \right],
\]
and

\begin{equation} \label{compReg_mod2}
\begin{array}{c}
\log\lambda_1(\bfs) = \omega^{\lambda_1}_{0} + \omega^{\lambda_1}_{1} \text{Elev}(\bfs) + \omega^{\lambda_1}_{2} \text{Dist2Coast}(\bfs), \\[0.5ex]
\log\lambda_2(\bfs) = \omega^{\lambda_2}_{0} + \omega^{\lambda_2}_{1} \text{Elev}(\bfs) + \omega^{\lambda_2}_{2} \text{Dist2Coast}(\bfs), \\[0.5ex]
\log\frac{\frac{2}{\pi}\gamma(\bfs)}{1 - \frac{2}{\pi}\gamma(\bfs)} = \omega^{\gamma}_{0} + \omega^{\gamma}_{1} \text{Elev}(\bfs) + \omega^{\gamma}_{2} \text{Dist2Coast}(\bfs).
\end{array}
\end{equation}
These specific transformations are chosen so that the coefficients $\{ \omega_j^{(\cdot)}\}$ can be unrestricted and still yield valid $\bfSig(\cdot)$.
For the sparsity-inducing kernel, the primary choice is to specify $n_1$ and $n_2$, the number of sets ($n_1$) of correlated yet distance-unrelated regions ($n_2$). Following the guidance of Section~\ref{sec:sde1}, we err on the side of choosing these to be larger than we might expect. We consider three configurations, each with 100 total bump functions: (1) $n_1 = 2$ and $n_2=50$; (2) $n_1=4$ and $n_2 = 25$; and (3) $n_1 = 10$ and $n_2 = 10$. Note that in this implementation, for the sparse kernel, we use an anisotropic version of the Wendland kernel that has separate $x$- and $y$-coordinate truncation radii. Furthermore, we consider three upper limits on the Wendland truncation radii (5, 10, and 15 units of degrees longitude/latitude) since we found that otherwise the truncation radii tend to increase without bound in the training. Limiting these radii to be relatively small ensures that the resulting kernel has the chance to impose sparsity on the data.

In summary, we have five hyperparameters for the prior mean function ($\{\alpha_j: j=0,\dots,4\}$), twelve hyperparameters for the core kernel ($\bftheta_\text{core} = \{ \phi_j, \omega_j^{\lambda_1}, \omega_j^{\lambda_2},\omega_j^{\gamma}: j = 0, 1, 2\}$), one hyperparameter for the noise variance. In the sparse kernel, 100 bump functions means that we have $3 + n_1n_2(d+3) = 3 + 100\times5 = 503$ hyperparameters. Overall, this yields $5 + 12 + 1+503 = 521$ hyperparameters for the GP applied to each time slice.


\subsection{Spatio-temporal modeling} \label{apdx:spattimemod}


\noindent \textbf{Prior mean function in space and time.} 
In light of the fact that we are modeling high-dimensional space-time data and our kernel can discover and capitalize upon sparsity, it is to our benefit to propose a prior mean function that explains as much of the known spatial and temporal structure in the data as possible. The most prominent structure in a data set involving temperature is the annual cycle, wherein daily maximum temperatures in the Northern Hemisphere are generally lowest in the winter and highest in the summer. Also, there is an underlying geospatial climatology to measurements of \newTxt{daily maximum} temperature including an implicit zonal structure wherein spatial locations farther from the equator are generally cooler, a strong influence from orographic variability, and the influence of coastlines. Specifically, we  expect (1) inverse relationships between temperature and elevation (via so-called ``lapse-rate'' theory); and (2) direct relationships between temperature and distance to coastlines. Elevation is obtained from the 800m digital elevation map used to generate the PRISM data product; slope and aspect are calculated using the \texttt{raster} package for \textbf{R} \citep{R_raster}. The distance-to-coast (km) variable is calculated by \citet{dist2coast} at a global grid of $1/16^\circ$ using the Generic Mapping Tools package \citep{wessel1998new} (note that this variable only calculates distances to ocean coastlines and not, e.g, other landlocked bodies of water). To account for each of these features, we propose the following prior mean function for each $\bfx = (\bfs,t) \in \mathcal{X} = \mathcal{S} \times \mathcal{T}$:
\begin{equation} \label{eq:priormean1}
\mu(\bfx) = \alpha_0({\bf s}) + \sum_{j=1}^J \alpha_j({\bf s})v_j(t).
\end{equation}
Here, the $v_j(\cdot)$ are natural cubic splines that map day $t$ to its Julian day to model the seasonal cycle, i.e.,
\[
v_j(t) \equiv \left\{
\begin{array}{ll}
     v_j\left(\cos\left([t \mod 365]\frac{2\pi}{365}\right)   \right) & j=1, \dots, J/2  \\
     v_j\left(\sin\left([t \mod 365]\frac{2\pi}{365}\right)   \right) & j = J/2+1, \dots, J
\end{array} \right.
\]
(swapping in 366 for 365 in leap years). Using sine- and cosine-based splines allows us to flexibly model the seasonal cycle such that the cycle in continuous in $t$; even a relatively low-dimensional representation (i.e., $J\approx 6$) explains roughly 75\% of the variability in the daily data.
Each of these cubic splines in time are assigned a spatially-varying coefficient, $\{ \alpha_j(\cdot): j = 0, \dots, J \}$, which we model using thin plate splines in the geographic coordinates, elevation, and distance-to-coast:
\begin{equation} \label{eq:priormean2}
\alpha_j({\bf s}) = \sum_{m=1}^{M_s} \delta_{jm}^s q^s_m({\bf s}) + \sum_{m=1}^{M_e} \delta_{jm}^e q^e_m(e)  + \sum_{m=1}^{M_c} \delta_{jm}^d q^c_m(c).
\end{equation}
Here, $\{ q^s_m(\cdot)\}$, $\{ q^e_m(\cdot)\}$, and $\{ q^c_m(\cdot)\}$ are sets of thin plate spline basis functions that define nonlinear mappings from the input coordinates ${\bf s}$, elevation $e \equiv e(\bfs)$, and distance-to-coast $c = c(\bfs)$, respectively. The thin plate spline basis functions are considered fixed  but their coefficients $\{\delta^s_{jm}\}$, $\{\delta^e_{jm}\}$, and $\{\delta^c_{jm}\}$ are considered hyperparameters and updated in the MCMC.

The linearity assumptions of Equations~\ref{eq:priormean1} and \ref{eq:priormean2} allow us to write the mean function $\mu(\bfx)$ as a linear function of a set of covariates, as denoted in Section~\ref{subsec:CGPM}. Let ${\bf W}_\text{time}$ denote a $N_\text{time} \times (J+1)$ matrix of cubic spline functions ($J$ of which correspond to the cubic splines and 1 for an intercept), where $N_\text{time} = |\mathcal{T}|$, and let ${\bf W}_\text{space}$ denote a $N_\text{space} \times (1+M_s + M_e + M_c)$ matrix of thin plate spline functions with an intercept, where $N_\text{space} = |\mathcal{S}|$. The design matrix ${\bf W}$ can then be written as
\begin{equation} \label{eq:W}
{\bf W} = {\bf W}_\text{space} \otimes {\bf W}_\text{time}
\end{equation}
(where $\otimes$ denotes the Kronecker product), which is applicable when the observational vector ${\bf z}$ concatenates all daily measurements from each station in turn, i.e., 
\[
{\bf z}=\big(z(\bfs_1, t_1), z(\bfs_1, t_2), \dots, z(\bfs_1, t_{N_\text{time}}), z(\bfs_2, t_{1}), \dots, z(\bfs_{N_\text{space}}, t_{N_\text{time}}) \big).
\]
(Here we suppose $N=N_\text{space}N_\text{time}$, although any inputs ${\bf x}$ with missing \newTxt{daily maximum} temperature measurements are omitted.) The matrix ${\bf W}$ in Equation~\ref{eq:W} is then $N\times [(J+1)\times(1+M_s + M_e + M_c)]$, meaning that this yields $(J+1)(1+M_s + M_e + M_c)$ hyperparameters for the prior mean. 
We re-label the coefficients $\{\delta^{(\cdot)}_{jm}\}$ from Equation~\ref{eq:priormean2} into an appropriately ordered vector $\bfbeta$ such that the prior mean function across all $N$ measurements can be concisely written as ${\bf W}\bfbeta$, as used in Section~\ref{subsec:CGPM}.

\begin{figure}[!t]
\begin{center}
\includegraphics[trim={0 0 0 0mm}, clip, width = \textwidth]{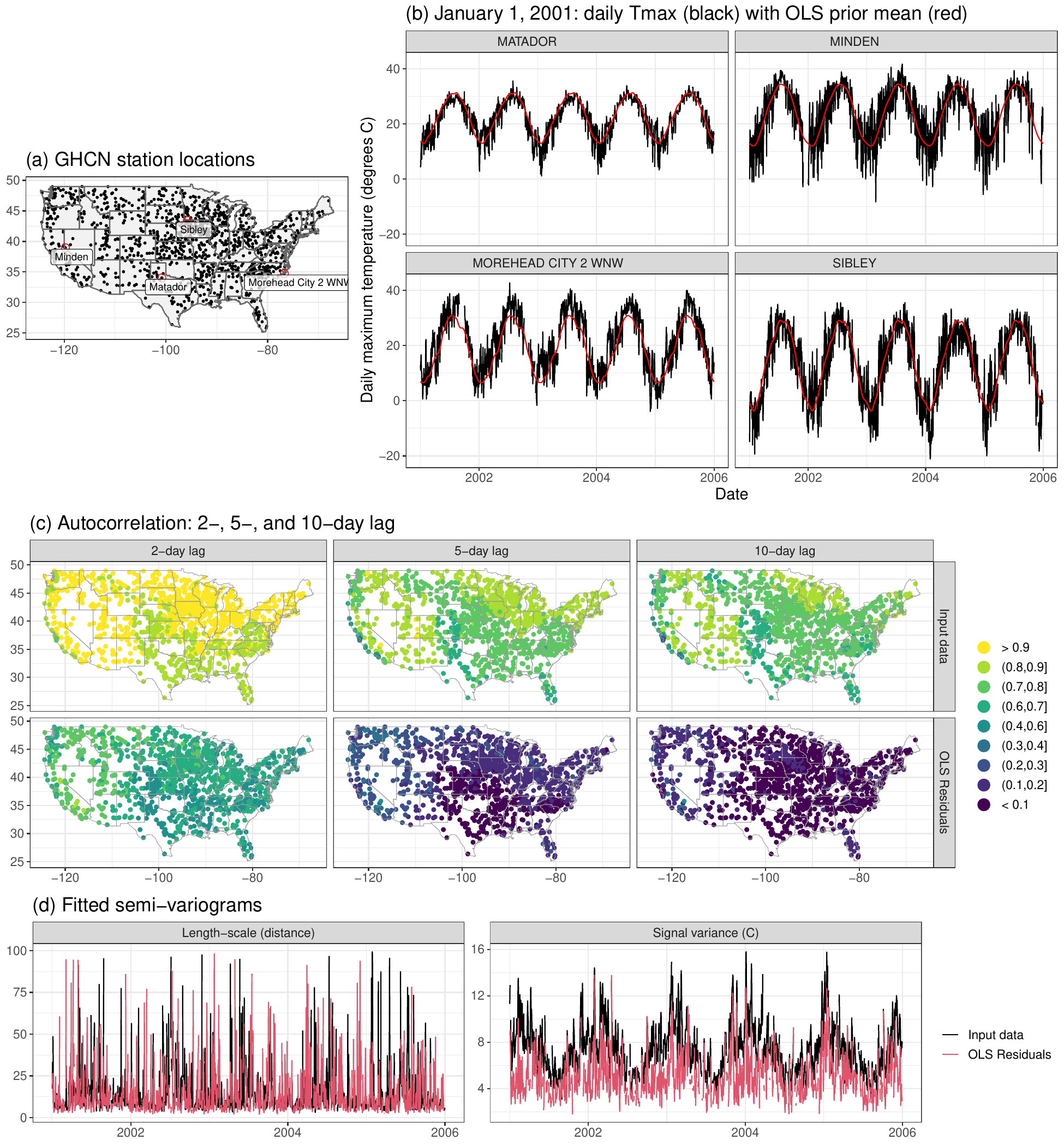}
\caption{Comparison of the fitted prior mean (using ordinary least squares) versus input data for four selected stations (shown in panel a.) across all time points (panel b.), where the raw data are shown in black and the fitted prior mean in red. Panel c. then shows spatial maps of the 2-, 5-, and 10-day lagged autocorrelation from the input data (top) and residuals (data minus OLS prior mean; bottom), while panel d. shows the length-scale and signal variance from semivariograms fitted separately for each for the input data (top) and residuals (data minus OLS prior mean; bottom).}
\label{Fig_priormean}
\end{center}
\end{figure}

\st{After some trial and error and i} \newTxt{I}n order to maintain a compromise between sufficient flexibility and overfitting, we fix $J=8$, $M_s=200$, $M_e = 5$, and $M_c = 5$, resulting in $1899$ hyperparameters for the prior mean. 
To visualize the selected prior mean function, we show the fitted prior mean -- for starters using ordinary least squares (OLS), but later these hypereparameters will be trained online with the GP -- for specific stations and days versus the corresponding raw temperature measurements; see Figure~\ref{Fig_priormean}. Across all input locations, the OLS linear mean explains 94.1\% of the variability in the daily temperature data, which is impressive since there is one hyperparameter for every $\approx 1100$ data points. For the specific stations shown in Figure~\ref{Fig_priormean}(b), it is clear that the seasonal cycle appropriately captures the general trend of the data, such that the spatially-varying nature of the coeffients can describe different amplitudes and even shapes in the the seasonal cycle: note that the daily maximum temperatures range from approximately 5$^\circ$C to 35$^\circ$C at the Matador station and approximately -10$^\circ$C to 30$^\circ$C at the Sibley station; furthermore, the seasonal cycle is extremely smooth at the Minden station but ``bumpy'' at Morehead City. 

It is important to note how the variability explained by a flexible prior mean function ``increases'' the sparsity that can be leveraged by the kernel. In the time dimension, Figure~\ref{Fig_priormean}(c) shows how the 2-, 5-, and 10-day lagged autocorrelations at each gauged location differ for the input data versus the residuals (the input data minus the OLS prior mean function). Even at a lag of 10 days, the data show strong autocorrelations, in excess of 0.6. However, when removing the fitted OLS prior mean, 10-day lagged autocorrelations essentially vanish; even for 5-day lags, the autocorrelations are quite small. Spatially, Figure~\ref{Fig_priormean}(d) shows the length-scale and signal variance from a variogram fitted to either the input data (black) or the OLS residuals separately for each day. Here, the effect of the prior mean function is less clear: in particular, the length-scale of the residuals appear to be on a similar order of magnitude as the input data. On the other hand, the variability of the residuals is significantly reduced relative to the input data.
In summary, it is clear that the prior mean function allows the data to be quite sparse in the time domain (i.e., correlations beyond $\pm5$ days can be safely ignored) and perhaps less sparse in the spatial domain.
These differences in the structure explained by the prior mean (temporal vs. spatial) have important implications for how we set up the kernel.

\vskip2ex
\noindent \textbf{Kernel.} For the full space-time data, we use a core kernel similar to the one in Equation~\ref{eq:cov_timeslice}, albeit generalized to a three-dimensional input space, i.e., $C_\text{core}(\bfx, \bfx')$. The anisotropy matrix again uses the eigendecomposition $\bfSig(\bfx)={\bf \Gamma}(\bfx) {\bf \Lambda}(\bfx) {\bf \Gamma}(\bfx)^\top$, where
\[
{\bf \Lambda}(\bfx) = \left[ \begin{array}{ccc} \lambda_1(\bfs) & 0 & 0 \\ 0 & \lambda_2(\bfs) & 0 \\ 0 & 0 & \lambda_3(t) \end{array} \right], \hskip3ex
{\bf \Gamma}(\bfx) = \left[ \begin{array}{ccc} \cos \gamma(\bfs) & -\sin \gamma(\bfs) & 0 \\ \sin \gamma(\bfs) & \cos \gamma(\bfs) & 0 \\
0 & 0 & 1 \end{array} \right].
\]
The formulas for $\lambda_1(\bfs)$, $\lambda_2(\bfs)$, and $\gamma(\bfs)$ are as in Equation~\ref{compReg_mod2}, and the length-scale in time is a constant, i.e., $\lambda_3(t) \equiv \lambda_3$. Note that in this case, the term inside the Mat\'ern correlation is 
\[
\sqrt{(\bfx - \bfx')^\top \left(\frac{\bfSig(\bfx) + \bfSig(\bfx')}{2}\right)^{-1}(\bfx - \bfx')} =  \sqrt{(\bfs - \bfs')^\top \left(\frac{\bfSig(\bfs) + \bfSig(\bfs')}{2}\right)^{-1}(\bfs - \bfs') + \frac{(t-t')^2}{\lambda_3}},
\]
and, since
\[
{\bf \Sigma}(\bfx) = \left[ \begin{array}{cc} \bfSig(\bfs) & {\bf 0} \\ {\bf 0} &  \lambda_3 \end{array} \right],
\]
we have
\[
\frac{\left|\bfSig(\bfx)\right|^{1/4}\left|\bfSig(\bfx')\right|^{1/4}}{\left|\frac{\bfSig(\bfx') + \bfSig(\bfx')}{2} \right|^{1/2}} = \frac{\left|\bfSig(\bfs)\right|^{1/4} \lambda_3^{1/4} \left|\bfSig(\bfs')\right|^{1/4} \lambda_3^{1/4}}{\left|\frac{\bfSig(\bfs') + \bfSig(\bfs')}{2} \right|^{1/2}\lambda_3^{1/2}} = \frac{\left|\bfSig(\bfs)\right|^{1/4}\left|\bfSig(\bfs')\right|^{1/4}}{\left|\frac{\bfSig(\bfs') + \bfSig(\bfs')}{2} \right|^{1/2}}.
\]
Next, the signal variance is again a log-linear function
\[
\log \sigma(\bfx) = \phi_{0} + \phi_{1} \text{Elev}(\bfs) + \phi_{2} \text{Dist2Coast}(\bfs), 
\]
which focuses on spatial heterogeneity in the signal variance. 
It is important to note that this specification yields a non-separable space-time kernel, i.e., 
\[
C_\text{core}(\bfx, \bfx') \neq C(\bfs, \bfs') \times C(t, t').
\]
For the sparse kernel, the Wendland component is again anisotropic with coordinate-specific length scales (and coordinate-specific truncation limits); following Section~\ref{sec:spatmod} we use 10 units for the spatial truncation limits and 5 days for the temporal truncation limits based on the fact that the prior mean function explains autocorrelation beyond 5 days for a large majority of the domain. We again use 100 bump functions with $n_1=4$ and $n_2=25$; however, again based on the fact that the prior mean explains so much of the temporal variability, we restrict the bump functions to live in the spatial domain only. In total, this yields $521 + 1 + 1= 523$ hyperparameters for the kernel (the same 521 from the \st{time-slice} \newTxt{spatial-only} modeling in Section~\ref{sec:spatmod} plus $\lambda_3$ and the Wendland length-scale in the time dimension).

\clearpage


\section{Supplemental figures}

\begin{figure}[!h]
\begin{center}
\includegraphics[trim={0 0 0 0mm}, clip, width = 0.7\textwidth]{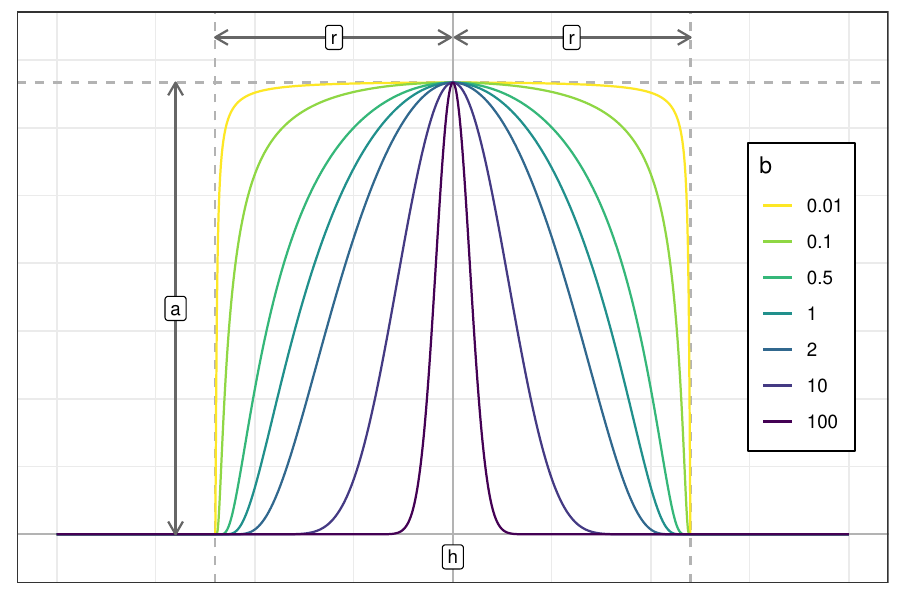}
\caption{Sample bump functions $g(\bfx)$ from Equation~\ref{eq:bumpFcn} for different values of the shape parameter $b$ with fixed amplitude $a$ and radius $r$. As $b\rightarrow0$, the bump function becomes uniform on $(-r, r)\subset \mathbb{R}^1$; as $b\rightarrow \infty$, the bump functions become delta function  centered at $\mathbf{h}$. Note that the bump functions are exactly zero for distances from the centroid ${\bf h}$ greater than or equal to $r$.}
\label{Fig_BumpFcns}
\end{center}
\end{figure}

\begin{figure}[!t]
\begin{center}
\includegraphics[trim={0 0 0 0mm}, clip, width = \textwidth]{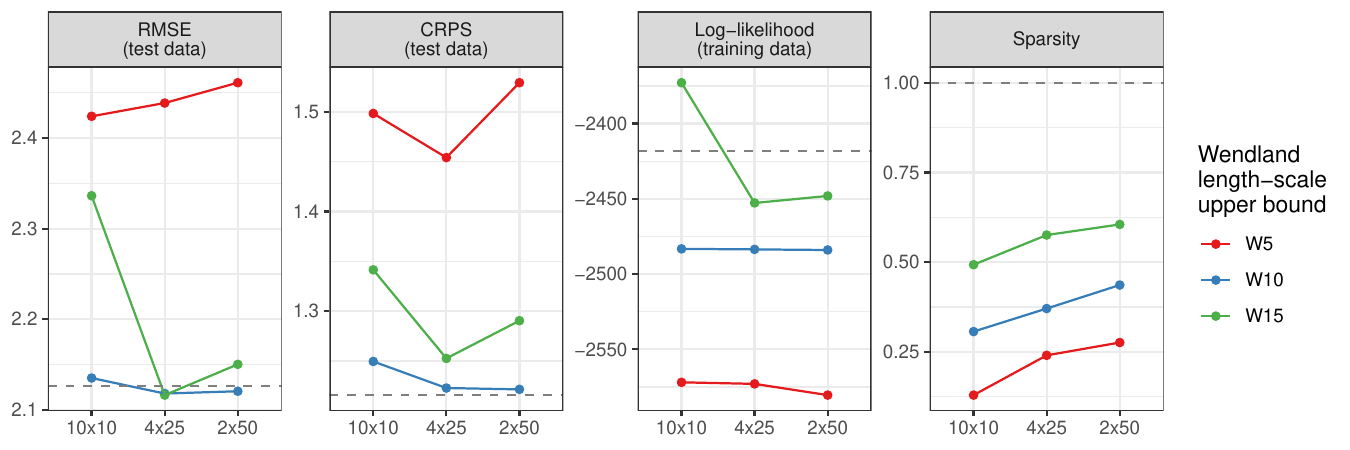}
\caption{Root mean square error (RMSE), continuous rank probability score (CRPS), log-likelihood, and sparsity (defined as the fraction of the covariance that is nonzero) from analyzing the January 14, 2001 time slice of daily maximum temperature. We show results for the three configurations of $n_1$ and $n_2$ as well as the three upper bound limits on the Wendland length scale (``W$x$'' indicates that the Wendland length scale is capped at $x$ units). The dashed gray line shows results from fitting a nonstationary Gaussian process with the same core kernel but without the sparsity-inducing kernel.
}
\label{Fig_timeslice0}
\end{center}
\end{figure}

\begin{figure}[!t]
\begin{center}
\includegraphics[trim={0 0 0 0mm}, clip, width = \textwidth]{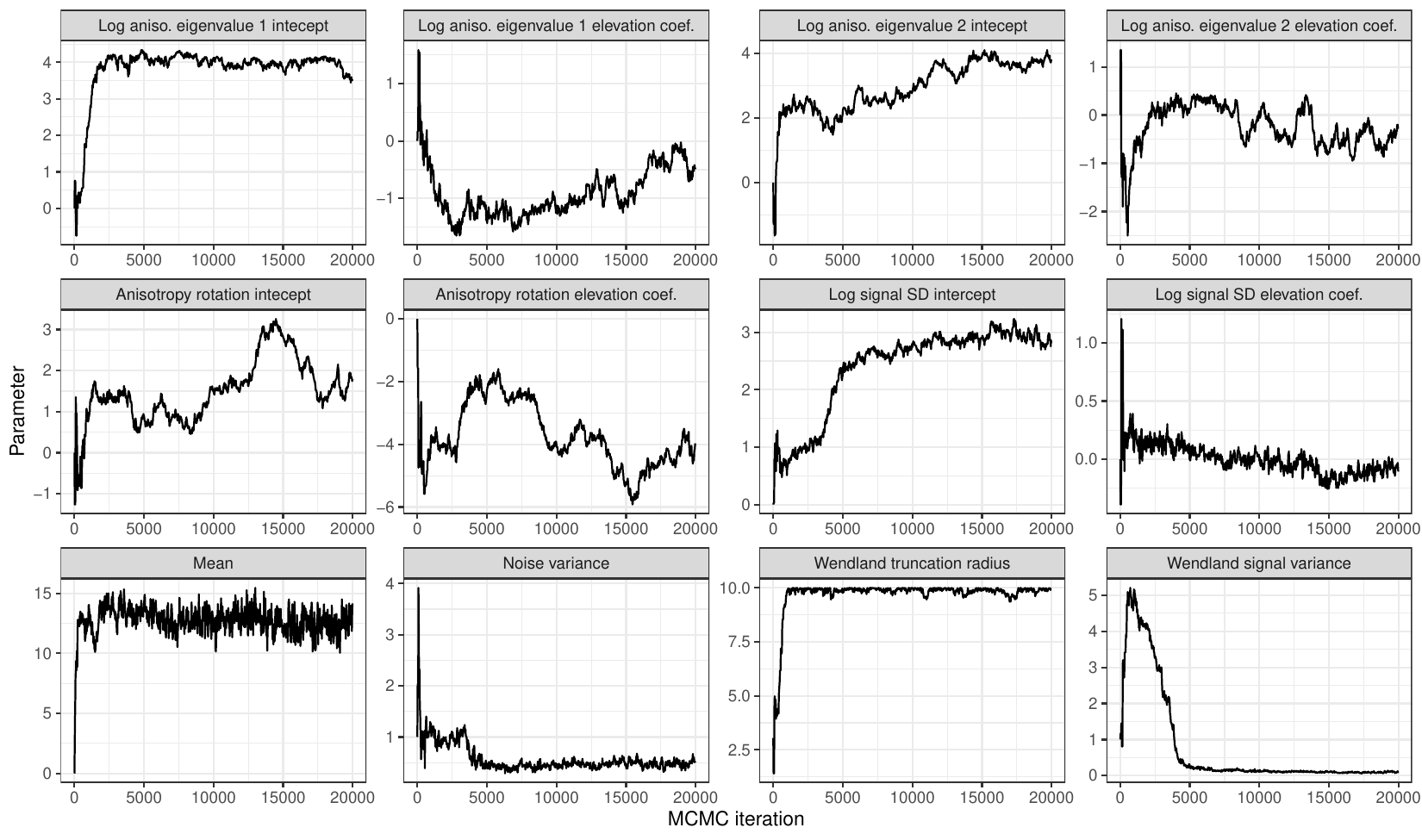}
\caption{Trace plots for the January 14, 2001 time slice of daily maximum temperature using the $n_1=4$, $n_2=25$ configuration.  
}
\label{Fig_trace1}
\end{center}
\end{figure}

\begin{figure}[!t]
\begin{center}
\includegraphics[trim={0 0 0 0mm}, clip, width = \textwidth]{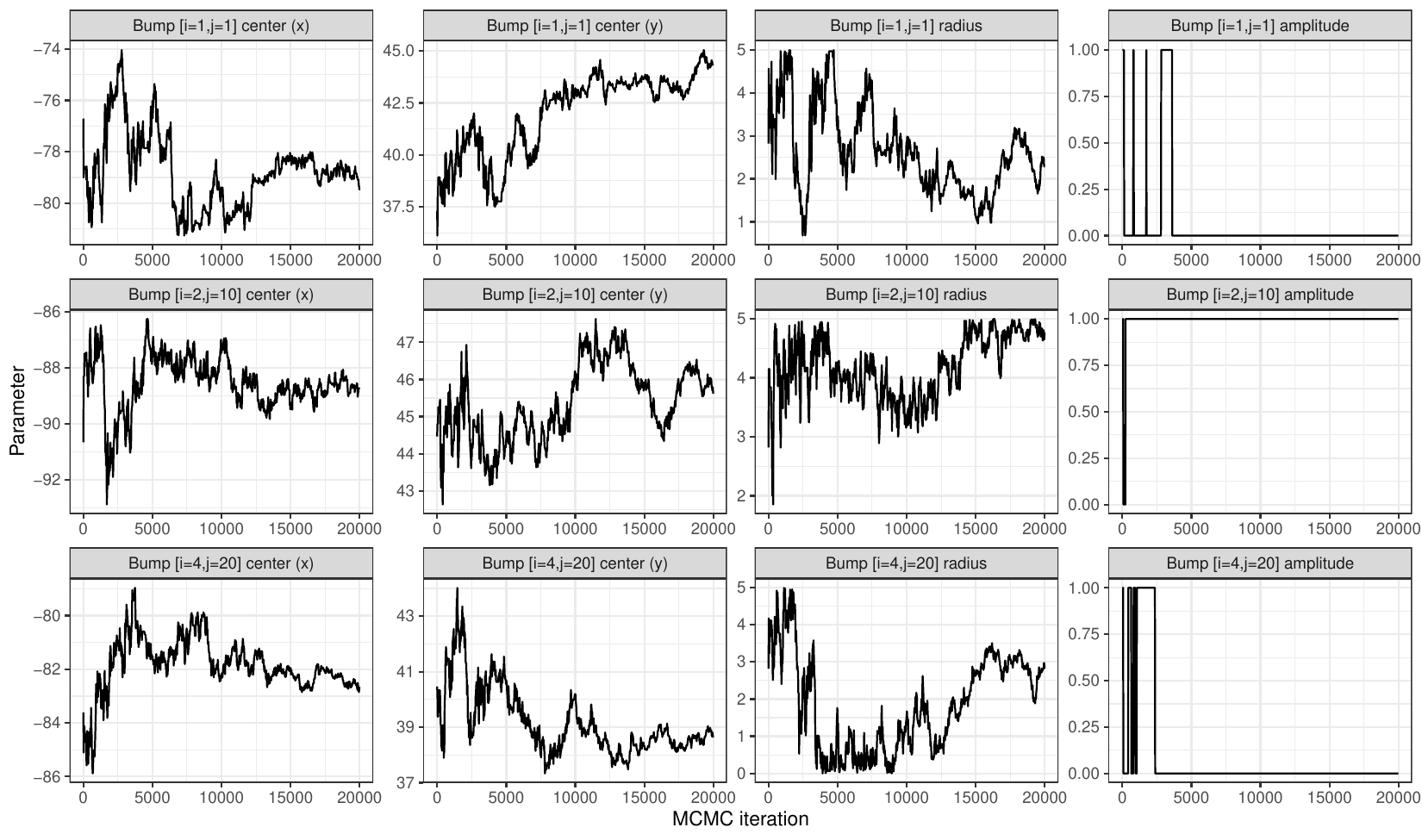}
\caption{Trace plots for the January 14, 2001 time slice of daily maximum temperature using the $n_1=4$, $n_2=25$ configuration.  
}
\label{Fig_trace2}
\end{center}
\end{figure}

\begin{figure}[!t]
\begin{center}
\includegraphics[trim={0 0 0 0mm}, clip, width = \textwidth]{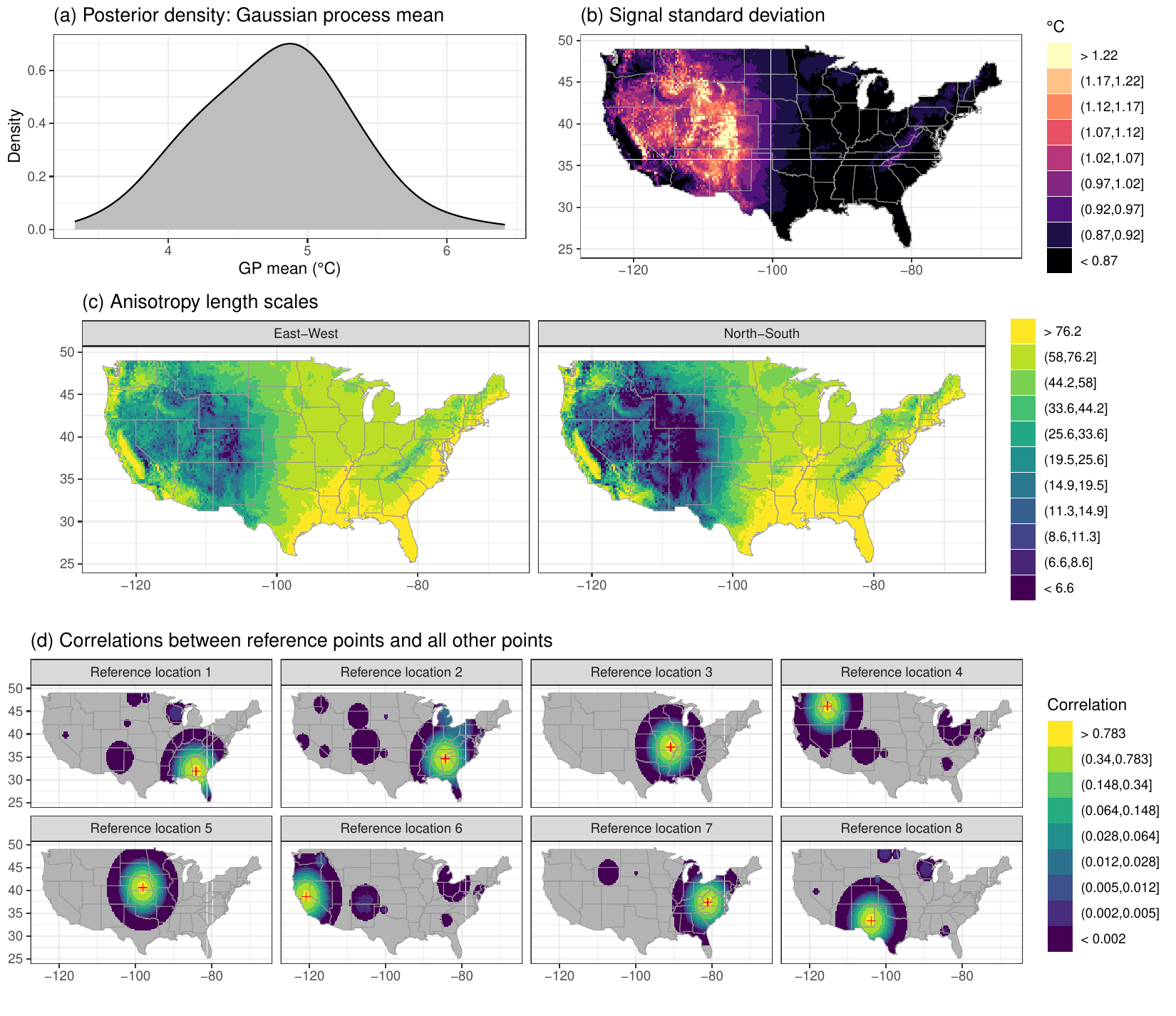}
\caption{\st{Posterior summaries of hyperparameters for the January 14, 2001 time slice of daily maximum temperature. Panels (a), (b), and (c) show the prior mean function, square root of the signal variance, and anisotropy length-scales, calculated using the posterior median hyperparameters and the physical covariates. Panel (d) shows implied correlations between eight reference locations and all other points in the input domain.}
\newTxt{Posterior summaries of hyperparameters for the January 14, 2001 time slice of daily maximum temperature. Panel (a) shows the posterior density of the prior mean; panels (b) and (c) show the square root of the signal variance and anisotropy length-scales, calculated using the posterior median hyperparameters and the physical covariates. Panel (d) shows implied correlations between eight reference locations and all other points in the input domain.}
}
\label{Fig_timeslice1}
\end{center}
\end{figure}

\begin{figure}[!t]
\begin{center}
\includegraphics[trim={0 0 0 0mm}, clip, width = \textwidth]{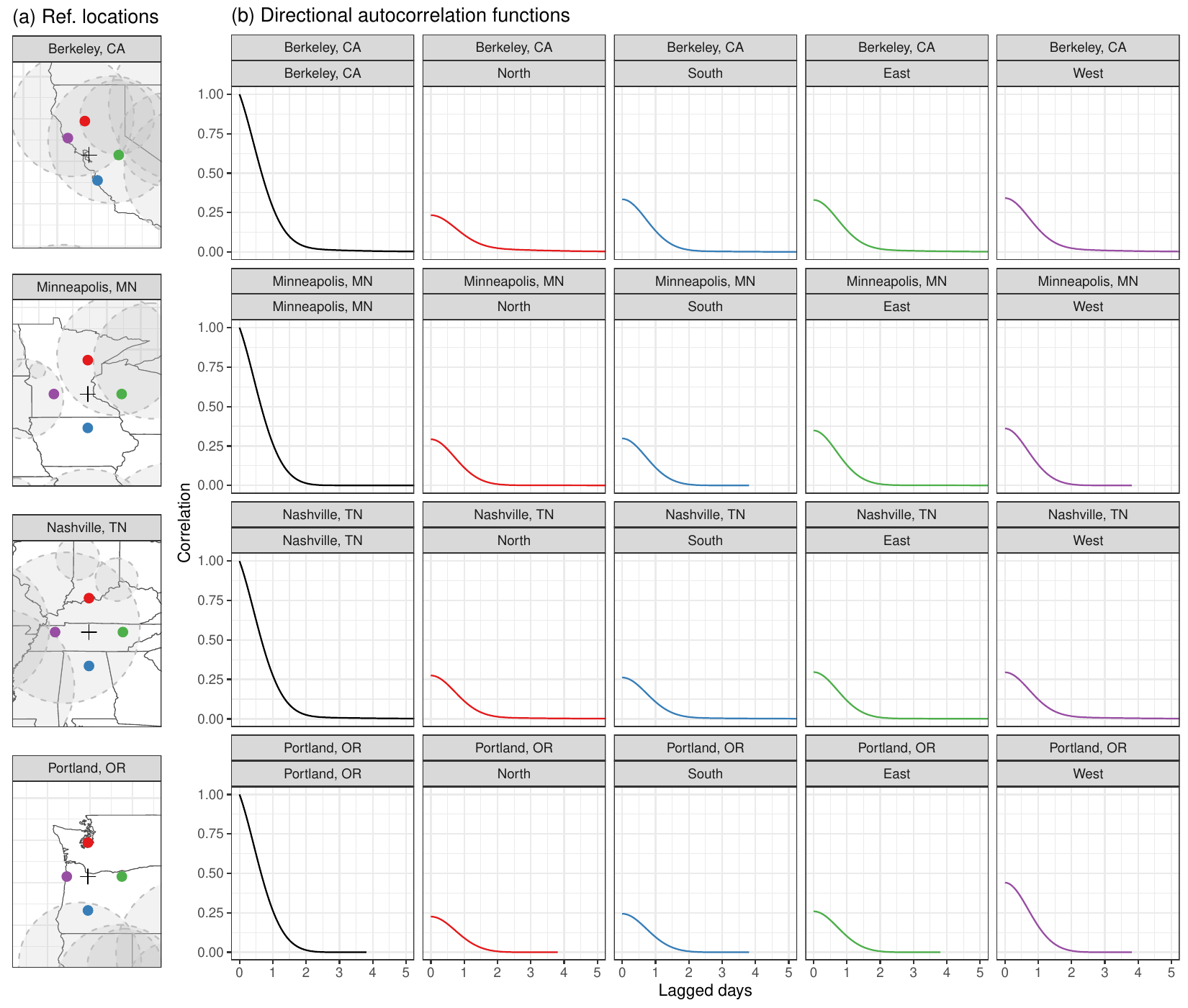}
\caption{Directional autocorrelation functions for pairs of locations over time. Here, we focus on four reference locations: Berkeley, CA; Minneapolis, MN; Nashville, TN; Portland, OR. For each of these reference locations, we compare correlation at four sites that are $2^\circ$ north, south, east, and west. In panel (a), the activated bump functions (i.e., those with $a_{ij}=1$) are shown in dashed gray. In panel (b), correlations are only plotted when they are nonzero. 
}
\label{Fig_corr_fcns2}
\end{center}
\end{figure}

\clearpage

\section{Supplemental tables}

\begin{table}[!h]
    \centering
\begin{tabular}{p{2cm}p{6cm}p{6cm}}
    \textbf{Parameter} & \textbf{Description} & \textbf{Default values}  \\\hline 
      $n_1$   & The number of bump function sums; defines the maximum possible groups of correlated, distance-unrelated regions. & As large as possible, but realistically between 5 and 10. We found that $n_1 \leq n_2$ resulted in better performance. \\[1ex]
      $n_2$   & The number of bump functions in each $f_i$; defines the maximum possible regions of distance-unrelated correlations. & As large as possible, but realistically between 10 and 50.  \\[1ex]
      $D_0$   & Prior upper bound for the compactly supported Wendland kernel. & 5-15\% of the maximum inter-point distance.  \\[1ex]
      $D_r$   & Prior upper bound for the bump function radii. & 2-5\% of the maximum inter-point distance. \\[1ex]
      $l_k, u_k$   & Prior lower and upper bound, respectively, for the bump function centroids in dimension $k=1,\dots, d$ & The minimum and maximum values present in the data set. \\[1ex]
    \end{tabular}
    \vskip2ex
    \caption{Summary table of all user-specified parameters for the sparse kernel $C_\text{sparse}$, along with recommended default values.}
    \label{tab:parameters}
\end{table}


\end{appendix}
\end{document}